\documentclass[lettersize,journal]{IEEEtran}
\pdfoutput=1 
\usepackage{array}
\usepackage{amsmath,amsfonts}
\usepackage{mathptmx}      
\usepackage{algorithmic}
\usepackage{algorithm}
\usepackage[caption=false,font=normalsize,labelfont=sf,textfont=sf]{subfig}
\usepackage{textcomp}
\usepackage{stfloats}
\usepackage{url}
\usepackage{verbatim}
\usepackage{graphicx}
\usepackage{caption}
\usepackage{cite}
\usepackage{amssymb}
\usepackage{amsthm}
\usepackage{latexsym}
\usepackage{url}

\usepackage{hyperref}
\usepackage{preamble/beimath}
\usepackage{preamble/textoolz}

\usepackage{comment}
\usepackage{xcolor}
\usepackage{orcidlink}

\newcommand{\cmt}[1] {{\color{black}#1}}
\newcommand{\ct}[1] {{\color{black}#1}}

\begin{document}
\nocite{sharma2023continuous}

\title{Multi-field Visualization: Trait design and trait-induced merge trees}
\author{Danhua Lei \orcidlink{0000-0002-6134-0258}, Jochen Jankowai\orcidlink{0000-0002-8324-550X}, Petar Hristov\orcidlink{0000-0002-1482-2529}, Hamish Carr\orcidlink{0000-0001-6739-0283}, Leif Denby\orcidlink{0000-0002-7611-9222}, Talha Bin Masood\orcidlink{0000-0001-5352-1086} and Ingrid Hotz\orcidlink{0000-0001-7285-0483}
}
\maketitle
\begin{abstract}
Feature level sets (FLS) have shown significant potential in the analysis of multi-field data by using traits defined in attribute space to specify features in the domain. In this work, we address key challenges in the practical use of FLS: trait design and feature selection for rendering. To simplify trait design, we propose a Cartesian decomposition of traits into simpler components, making the process more intuitive and computationally efficient. Additionally, we utilize dictionary learning results to automatically suggest point traits. To enhance feature selection, we introduce trait-induced merge trees (TIMTs), a generalization of merge trees for feature level sets, aimed at topologically analyzing tensor fields or general multi-variate data. The leaves in the TIMT represent areas in the input data that are closest to the defined trait, thereby most closely resembling the defined feature. This merge tree provides a hierarchy of features, enabling the querying of the most relevant and persistent features. Our method includes various query techniques for the tree, allowing the highlighting of different aspects. We demonstrate the cross-application capabilities of this approach through five case studies from different domains.

\end{abstract}


\section{Introduction}
\label{sec:introduction}
%

Visualization of simulation data of natural phenomena typically involves the representation of multiple interacting fields. Such, so-called multi-fields, can be described as a mapping of the domain into a higher-dimensional attribute space. 
Despite the widespread use of multiple fields, their visualizations are much less studied compared to those of scalar or vector fields.
In particular, there remains a need for intuitive, easy-to-configure, feature-based representations of multi-fields.

Recent advances have addressed these issues by adapting key scalar field visualization methods for multi-variate data. Examples include volume rendering using feature space representatives~\cite{Jankowai2020a} and the generalization of isosurfaces to feature level sets (FLS)~\cite{JankowaiHotz2019}. In this context, \emph{multi-variate features} are introduced as regions within the domain that are induced by \emph{traits} in the attribute space. FLSs are then level sets of the corresponding induced distance field generated with respect to the trait. In this setting, the multi-variate feature itself is given by the zero FLS, which in the bi-variate case corresponds to fiber surfaces specified by a fiber surface control polygon~\cite{fibersurfaces}. A major advantage of FLS is that even if the zero level set is empty, users can investigate how closely the field approaches the trait by rendering different level sets of the distance field. As with fiber surfaces, the trait geometry is user-defined and specifies the parameter configurations of interest.

Since their introduction, feature level sets have been successfully applied in various domains, including uncertainty visualization~\cite{Sane2021} and flow visualization~\cite{Nguyen2021a}. However, the success of this concept strongly depends on the careful selection of traits and the iso-levels to be rendered. This paper addresses these limitations by proposing methods to support the design of traits in the attribute space, as well as using trait-induced merge trees that provide automatic levels or transfer function specifications for rendering the domain.

\subsubsection*{Trait design} 
To simplify the specification of traits for FLS generation, we propose two methods.
The first method is based on the observation that traits in attribute space are typically not arbitrarily complex geometries, but are composed of simpler traits in lower dimensions. 
Usually, these are intervals or points in a single dimension, or polygons in two or three dimensions. Exploiting this observation leads to what we call \emph{Cartesian traits}, described in Section~\ref{sec:cartesian}. These traits simplify both the trait design process and the computation of the distance field.
In the second approach, we use attribute space clustering and sparse dictionary learning~\cite{Lei2024}, as described in Section~\ref{sec:point-trait-sparse}, to automatically propose simple point traits, referred to as \emph{atom-traits}. These point traits can then be logically combined using the concept of Cartesian traits.

\subsubsection*{Trait induced merge trees} 
Once the feature definition via traits is established, the next step is to specify isovalues for the levels of FLS. The main goal is to highlight regions whose parameters are close to the trait. A straightforward solution is to represent FLS for a range of isovalues, but this can lead to visual clutter. Conversely, using a single global isovalue could overemphasize some regions and overlook other areas of interest. Topological analysis, particularly the use of merge trees, addresses these concerns effectively, as introduced by Weber et al.~\cite{Weber2007a}. We build on this idea by introducing trait-induced merge trees (TIMTs).
They open up an array of topology-based simplification and query methods that can provide further insight into the underlying structure of the data.
To automatically extract local isovalues and capture the most significant features, we propose various feature selection strategies within TIMTs, such as crown features~\cite{Nilsson2022}. This approach makes features extracted through FLS browsable via an interface for interacting with the corresponding merge tree. Combining FLS with scalar field topology offers a straightforward method for topological analysis of multi-variate fields. The method is flexible regarding the chosen features of interest, represented by traits in attribute space.

\cmt{This paper is an extended version of a previously submitted paper to TopoInVis 2023~\cite{jankowai2023multi}. It has been significantly expanded by including new methods for trait specification, 
\ct{namely}, Cartesian traits and point traits based on dictionary learning, that simplify the design of complex traits and improve computational efficiency. New case studies demonstrate these traits in high-dimensional data analysis and vortex-reconnection data, showcasing the practicality of automatic trait suggestions compared to manual design. Additionally, a stability analysis of TIMT has been added in the 
Appendix:~\ref{appendix_A}.} 

\begin{figure*}
    \centering
    \includegraphics[width=\linewidth]{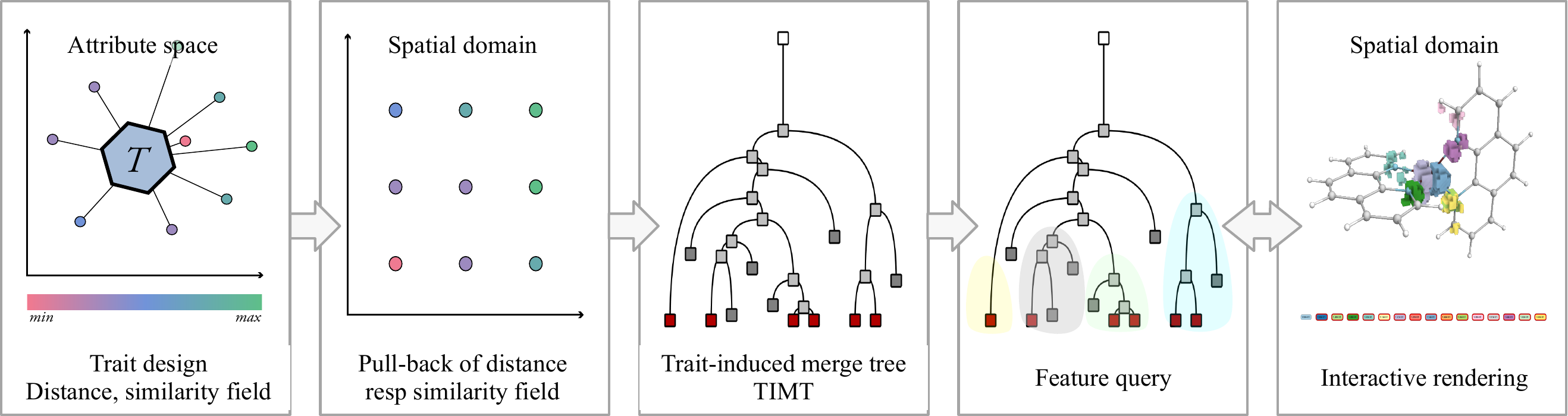}
    \caption{Pipeline. The first step is to design a trait representing the parameter settings of interest~\cmt{(Sec~\ref{Trait_design})}. For every vertex in attribute space, the distance resp similarity to the trait is calculated, which is then pulled back to the domain~(\cmt{Sec~\ref{FLS}}). The resulting distance field serves as input for the computation of the trait-induced merge tree (TIMT)~(\cmt{Sec~\ref{sec:TIMT}}) that then can be queried~(\cmt{Sec~\ref{sec:query_methods}}). Finally, the user can interact with the resulting domain segmentation via a legend or a slice of the data.}
    \label{fig:pipline}
\end{figure*}

The paper is structured according to the pipeline shown in~\fref{fig:pipline}.
After discussing related work in~\sref{sec:relatedwork} and providing some background in~\sref{sec:background} we focus on the trait design in~\sref{Trait_design}, specifying the parameter settings of interest. The trait-induced merge tree (TIMT) is introduced in~\sref{sec:TIMT}. \sref{sec:interaction} presents the different interaction and feature querying methods. Finally, we demonstrate the method in several case studies in~\sref{sec:results} before concluding with a discussion~\sref{sec:conclusions}. 
\section{Related work}
\label{sec:relatedwork}
Our work combines recent advancements in multi-field visualization and topological data analysis. In the following, we place our contributions in the context of existing research.
\subsection*{Attribute space interactions for multi-field visualization}
Coordinated linked views are frequently employed to visualize multi-fields, as highlighted in the state-of-the-art report by Roberts et al.~\cite{Roberts2007}. 
In much of this work, representations in attribute space play a crucial role in designing multi-dimensional transfer functions (TFs)~\cite{Ljung2016}.
This task is often supported by clustering or segmentation within the attribute space.
Wang et al.~\cite{Wang2012} propose segmenting a 2D density plot in attribute space using a Morse decomposition to generate a TF automatically.
Similarly, Cai et al.~\cite{Cai2017}  suggest a two-level approach that begins with topology-preserving dimensionality reduction, followed by a clustering step.
Dobrev et al.~\cite{Dobrev2011} introduce a method for interactive TF generation using a cluster hierarchy. Their method combines a cluster tree visualization with parallel coordinates to create an interactive interface.
Jankowai et al.~\cite{Jankowai2020a} present an interface that utilizes cluster representatives to design TFs for rendering tensor fields. 

%

%
\subsection*{Topology guided visualisation}
Concepts from topological data analysis, particularly the contour tree or merge tree are frequently used to guide visualizations. In an early paper, van Kreveld et al.~\cite{vanKreveld1997a} augmented a contour tree with seed sets to enable fast isocontour computation. Weber et al.~\cite{Weber2007a} presented an approach for volume rendering of topologically segmented scalar fields, assigning a distinct transfer function to each segment. 
Methods that integrate results from topological analysis into interactive frameworks have been especially successful. For example, Bremer et al.~\cite{Bremer2011} used a linked-view interface to analyze burning cells in turbulent combustion simulations. Bock et al.~\cite{Bock2018} employed a combination of merge tree analysis and an interactive user interface for the efficient segmentation of micro-CT scan data of fishes.
Besides the merge tree, other topological structures have also been used in visual frameworks. For example, Shivashankar et al.~\cite{Felix} introduced a queryable hierarchy of Morse-Smale complexes that allows astronomers to examine filamentary structures of the cosmic web at different scales. A comprehensive survey of topology-based methods in visualization can be found in the report by Heine et al.~\cite{topologystar}.

\subsection*{Level-set and topological concepts for multi-fields}
A core feature of the presented method is the rendering of relevant isosurfaces for multi-fields. 
Carr et al.~\cite{fibersurfaces} introduced the concept of fiber surfaces as a generalization of isosurfaces to bi-variate data using sets of fibers, which are the bi-variate equivalent to isolines. 
These allow for generating fiber surfaces based on control polygons (CP) in attribute space, \cmt{which can be considered as a set of line-traits in the context of FLS}. Since then, various extraction and rendering methods for fiber surfaces have been developed.
Wu et al.~\cite{Wu2017a} offered a system for the interactive exploration of bi-variate data through real-time pixel-perfect fiber surface rendering using intersection tests in range space on the fly. Klacansky et al.~\cite{Klacansky2017a} implemented a topology-agnostic and exact calculation of fiber surfaces significantly speeding up the process.
Fiber surfaces were later generalized to multi-variate data by Raith and Blecha et al. ~\cite{raithsalamislice1,raithsalamislice2}, who introduced interactors–user-defined geometries in attribute space. These geometries determine points in the spatial domain that form the isosurface.
Feature level sets (FLS)~\cite{JankowaiHotz2019} further extended multi-field isosurface visualization by generalizing isosurfaces to multi-variate data using distance field computation in attribute space with respect to traits. 
Nguyen et al.~\cite{Nguyen2021a}  and Athawale et al.~\cite{Athawale2021a}  applied FLS to separate and visualize structures in Taylor-Couette flow simulations and examine correlations in numerical simulations of solar farms, respectively.

\section{Data, traits, and merge trees}

\label{sec:background}
The proposed method is based on three main concepts: Multi-variate data, feature level sets and traits, and merge trees. A summary of these concepts is given below. 
%

%
\subsection{Multi-variate data}
We assume multi-variate data as input, given as a set of $m$ continuous fields $F_1, F_2, \cdots, F_m: \Xspace \to \Rspace$ defined on the \emph{data domain} $\Xspace$. 
We then construct an attribute space $\Aspace \subset \Rspace^M$ by combining selected field values and possibly some derived quantities. 
The multi-variate data is summarised by a multi-variate mapping, $f: \Xspace \to \Aspace$, where $M$ is the number of selected field values and their derived quantities.
Further, we assume that the attribute space $\Aspace$ is equipped with a metric (e.g., the Euclidean metric), denoted as $d_{\Aspace}$.

Typically, the data is given on a sampled discrete set of $N$ points in the domain $\Xspace$. For each point $x_{i}\in \Xspace, i=1\dots N$, the field values can be represented as a vector $f(x_{i})=[F_{1}(x_{i}), F_{2}(x_{i}),..., F_{m}(x_{i})]^{T}\in \Rspace^{M}$. 
These vectors are assembled into a matrix $\mathbf{F} \in \Rspace^{M \times N}$, where each column of $\mathbf{F}$ represents the field values at a specific point in the data domain.

%



%
\subsection{Feature level sets}
\label{FLS}
Feature level sets are built on the concept of \emph{trait-induced features}. Thereby, a \emph{trait} $T$ is defined as a subset in the attribute space, $T \subset \Aspace$. Examples of traits include convex polygons, points, collections of points, or line segments. A \emph{trait-induced feature} is the pre-image of a trait in the data domain, $f^{-1}(T) = \{x \in \Xspace \mid f(x) \in T \subset \Aspace\}$. Since the feature corresponding to an arbitrary trait may be empty, feature level sets have been introduced to highlight areas in the domain with values close to the trait. Therefore, a \emph{trait distance field}, $d_T: \Aspace \to \cmt{\Rspace^{+}}$, is defined, where $d_T(a) = \min_{t \in T} d_{\Aspace}(a, t)$ for $a \in \Aspace$. The \emph{trait-induced distance field} (or feature distance field) is a scalar function defined as $h_T = d_T \circ f: \Xspace \to \Rspace$.
Finally, the \emph{trait-induced level sets} are the level sets of $h_T$, given by $h_T^{-1}(c) = \{x \in \Xspace \mid h_T(x) = c\}$. In this paper, we consider alternatively \emph{trait similarity fields}, $s_T: \Aspace \to \Rspace$, where $s_T(a) = \max_{t \in T} d_{\Aspace}(a, t)$ for $a \in \Aspace$.

\subsection{Merge trees}
Let  $g: \Xspace \to \Rspace$ be a continuous scalar field. For the computational purpose, assume $g$ is defined on a simply connected compact simplicial complex $\Xspace$ and is linearly interpolated on the interiors of its simplices.
Two points $x, y \in \Xspace$ are considered \emph{equivalent}, denoted by $x \sim y$, if $g(x) = g(y)$ and $x$ and $y$ are a part of the same connected component of the sub-level set $g^{-1}((-\infty,g(x)])$. 
The quotient space $\Xspace/\sim$ is called a \emph{merge tree} of $g$. The merge tree records birth, death, and merge events of sub-level set components during a sweep of $g$ from $-\infty$ to $\infty$.
Typically, merge trees are computed using algorithms based on the work by Carr et al.~\cite{Carr2003a}. It is based on a \emph{sub-level set filtration} of $g$, observing changes in a sequence of nested sub-level sets connected by inclusions. Analogously, a definition for super-level sets $g^{-1}([g(x), \infty))$ can be formulated.

\subsection{Merge tree simplification}
\label{sec:simplification}
Since the full merge tree may contain many leaves, hierarchical representations providing a multi-scale view of the data are often used. Two common metrics considered in our pipeline for building such a hierarchy are persistence and hypervolume.

\begin{figure}
    \centering
    \subfloat[\centering]
    {
    \includegraphics[width=0.43\linewidth]{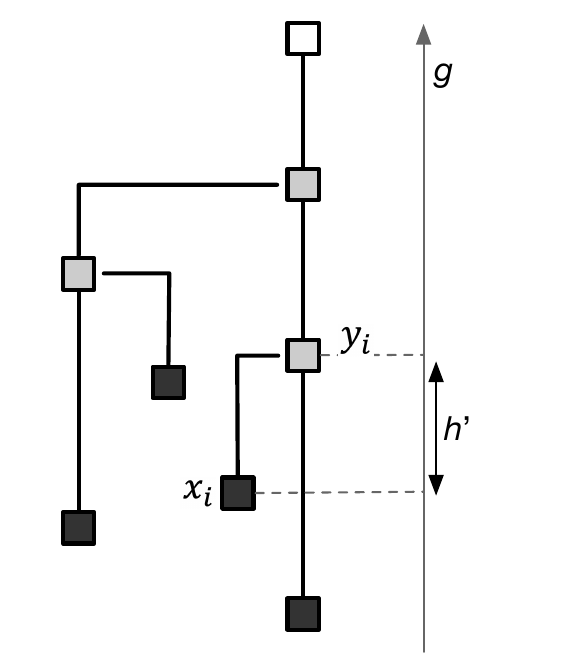}
    \label{subfig:metric_persistence}} 
    \subfloat[\centering]
    {
    \includegraphics[width=0.43\linewidth]{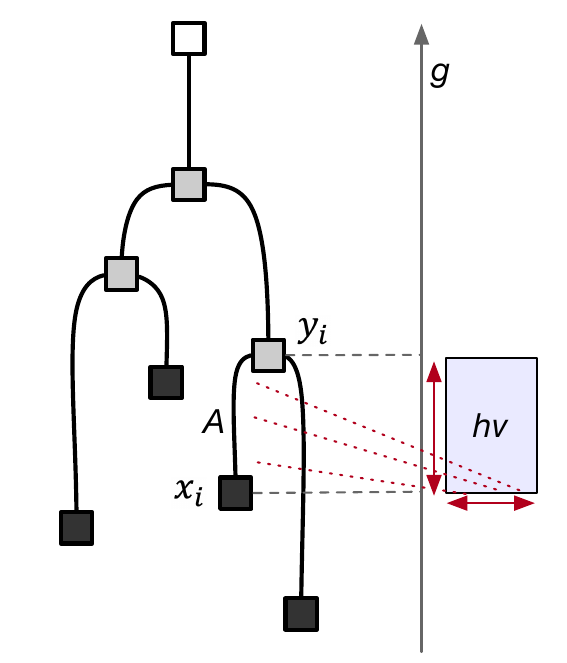}
    \label{subfig:metric_hypervolume}}
    \caption{Simplification metrics: \protect\ref{subfig:metric_persistence} The branch decomposition tree highlights the persistence of paired critical points $y_i$ and $x_i$, defined as the difference in their function values, $h' = g(y_i) - g(x_i)$. \protect\ref{subfig:metric_hypervolume} Hypervolume is calculated by accumulating the volume associated with an arc of the tree multiplied by its height. While persistence relates only to the data range, hypervolume considers also the spatial embedding of the data.}
    \label{fig:metric}
\end{figure}

\paragraph{Persistence}
In context with merge trees, persistence arises from a \emph{sub-level set filtration} generating a pairing of critical points $(x_i, y_i)$. During filtration, a feature is generated in one critical point and disappears in the other. The interval spanned by the function values of the critical points $[g(x_i),g(y_i)]$ represents the feature lifetime interval. To each pair of critical points, one can then assign a \emph{ persistence value} which is the difference of the scalar values in the two critical points $g(x_i)-g(y_i)$, see \fref{subfig:metric_persistence}. It gives some notion \cmt{of} feature stability~\cite{Cohen-SteinerEdelsbrunnerHarer2010}. This critical point pairing can be used for a controlled simplification of the data removing features ordered by their persistence value.
%
A \emph{branch decomposition tree} derived from a merge tree is a hierarchical representation of these pairs of critical points~\cite{Pascucci2005a}, see \fref{fig:query}(b).

The geometric interpretation of a low persistence feature in a two-dimensional example would be a shallow valley on a height-field map. Likewise, a high persistence value equates to a deep intrusion in a field. Low persistence values often occur in noise where small differences in function values create irrelevant extremal points. 
\paragraph{Hypervolume}
Hypervolume ~\cite{Bock2018} is a measure taking the local geometric extent of the sub-level sets in the data domain into account. As illustrated in \fref{subfig:metric_hypervolume}, hypervolume is the product of arc height and the volume contained by the region corresponding to the arc. In our implementation, this is the number of voxels contained in a segment multiplied by the difference in function values at the minimum and the saddle connected by the arc. 
%

%

%

\section{Trait design}
\label{Trait_design}

Trait specification is a crucial step in the pipeline, requiring a thorough understanding of the data. Our framework supports several basic methods for manually constructing traits. 
To enhance this process, we offer automatic suggestions for point traits based on data-specific dictionaries, as well as a Cartesian combination of basic traits that we call \emph{Cartesian traits}.

\subsection{Interactive trait specification}
 \begin{figure}[h]
     \centering
     \includegraphics[width=\linewidth]{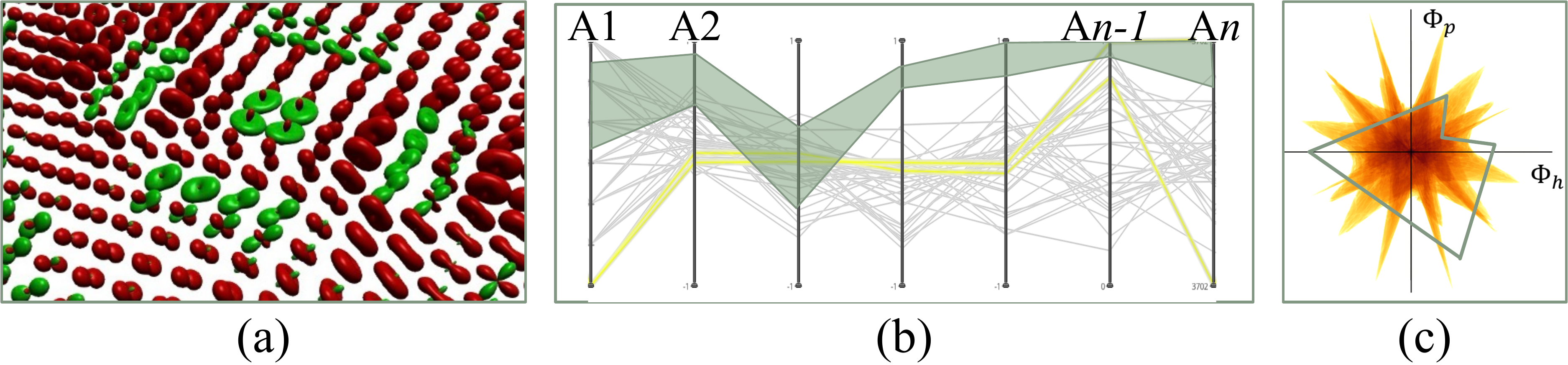}
     \caption{
    Methods for manually specifying traits: (a) selecting point traits by choosing a glyph from a rendered tensor slice, (b) defining traits as lines or areas in parallel coordinates, and (c) drawing polygons within a bi-variate subspace. 
     }
     \label{fig:manual}
 \end{figure}
Similar to the original paper on feature level sets~\cite{JankowaiHotz2019}, we offer an interface for manually specifying traits. This interface allows users to configure attribute spaces using available parameters selected from a drop-down menu. The configured attribute space is visualized in parallel coordinates, enabling the simultaneous representation of all parameters. Alternatively, two variables can be chosen to form a bi-variate subspace. Both visualizations facilitate trait specification and provide an overview of the data, such as a continuous scatterplot in the bi-variate setting.
This setup enables the specification of point traits as lines in the parallel coordinates plot or cubical traits as a set of intervals,~\fref{fig:manual}(b) or more general polygons in the bi-variate setting,~\fref{fig:manual}(c). 
For tensor fields, \cmt{where tensors are typically visualized as domain-specific glyphs} our interface also supports the selection of tensors as point traits through a glyph rendering of the dataset,~\fref{fig:manual}(a).

\subsection{Combinations using Cartesian traits}
\label{sec:cartesian}

For more advanced manual trait specification, we support the combination of multiple traits via \emph{Cartesian traits}, an extension of Cartesian fiber surfaces~\cite{hristov_thesis_2022}. The idea behind Cartesian traits stems from the observation that scientists rarely conceptualize features in a high-dimensional space. Thus, it is more intuitive to define high-dimensional traits implicitly as combinations of simpler low-dimensional traits, using robust trait primitives such as points, lines, polygons, and polyhedrons.
For example, instead of defining a trait in four-dimensional space directly, we decompose it as the region where the scalar $v_1$ is in the range of values $[a, b]$ \textit{and} the scalars $v_2, v_3$ are \textit{not} in an area $A$ of their joint distribution \textit{or} the scalar $v_4$ is \cmt{below} a certain threshold $c$.

We can combine two traits $T_1 \subset \Aspace_1 $ and $T_2 \subset \Aspace_2$ using the boolean operators AND, OR, and NOT to obtain a higher dimension trait $T$ in the Cartesian product space $T \subset \Aspace_1 \times \Aspace_2$.
For example, to obtain all points inside the volume bounded by the feature level set 
$h_{T_{1}}^{-1}(c_1)$ of $T_1$ and inside the volume bounded by the feature level set $h_{T_{2}}^{-1}(c_2)$ of $T_2$, we can combine the trait-induced distance fields with efficient distance field operations~\cite{frisken_2006}.
We can use the min operator and define the distance field of the combination of the traits as $h_{(T_1 \text{ \& } T_2)}(x) = min(h_{T_{1}}(x), h_{T_{2}}(x))$ for all $x \in \Xspace$.
The boolean operations OR and NOT are obtained using the max operator and negation, respectively.
This approach can be extended to combine any number of traits with any of the boolean operators.

\subsection{Atom-traits using Dictionary learning}
\label{sec:point-trait-sparse}

We propose using dimensionality reduction methods, such as data-specific dictionaries from sparse representation, to provide automatic guidance for trait design. This can be effectively achieved through dictionary learning~\cite{Lei2024}.

\cmt{Dictionary learning decomposes data into an overcomplete set of basis functions~\cite{elad2010sparse}, where the number of basis functions (or atoms) exceeds the dimensionality of the data. The main idea is to cluster data into distinct categories, each represented by a linear combination of selected atoms. \ct{Coefficients for unselected atoms remain zero, which creates a sparse representation.}  By using a minimal subset of atoms for each category, this approach provides a compact, interpretable representation that captures essential data features and patterns.}

Given a discrete set of $N$ points $\mathbf{x}$ in the data domain $\Xspace$ and their corresponding values $F_{j}(\mathbf{x})\,(j=1,2,\ldots, M)$ across $M$ different fields, this multi-variant data is typically represented as a matrix $\mathbf{F}\in\mathbb{R}^{M\times N}$ in the context of sparse representation. 
\cmt{The goal is to find a dictionary $D \in\mathbb{R}^{\cmt{M} \times K}$, which contains $K$ column vectors or atoms, and its corresponding sparse coefficients $C\in\mathbb{R}^{K\times \cmt{N} }$ to represent the data in the form of matrix multiplication \cmt{$\mathbf{F}=DC$}. The number of atoms $K$ is generally higher than the dimension of the original attribute space $M$. The sparsity level determines the maximum number of atoms used to represent each data point.}
This task can be formulated as the following optimization problem
\begin{equation}    \label{eq:ksvd}
    \min\limits_{D,C} \;\ \lVert \mathbf{F}-DC\rVert_{F}^2 
    \;\;\; s.t.\;\;\; \|C_{i}\|_{0} \le T_{0}, \; \forall i\in\{1,\dots,\cmt{N}\}, 
\end{equation}
where $\|\cdot\|_{F}$ denotes the Frobenius norm, $T_0$ is a user-defined parameter to control the sparsity level. For details on obtaining the dictionary and its coefficients using alternative optimization of equation \ref{eq:ksvd}, we refer readers to Section 4 of~\cite{Lei2024}.

This approach can loosely be interpreted as a 'weak clustering', where each data point is associated with a few clusters represented by the atoms with non-zero coefficients. This interpretation suggests that the atoms serve as a set of basis functions that effectively represent the data. This leads to the idea of using atoms as point traits, referred to as \emph{atom-traits} to explore features.
To assess how well atoms align with the data points, we utilize cosine similarity as a measure. This metric quantifies how closely two vectors align in multidimensional space, focusing on their directional similarity irrespective of magnitude. A value closer to 1 indicates higher similarity in the multidimensional space.
Using coefficients, we can explore any region within the volume to determine which atoms are involved and their contribution to data or feature representation. Furthermore, we provide support for users to use logical operators such as AND and OR to visualize combinations of different traits,~\sref{sec:cartesian}.

\section{Trait-induced merge trees}
\label{sec:TIMT}
Feature level sets (FLS) allow users to extract features as isosurfaces, highlighting areas in the domain with values close to the selected trait. However, \cmt{as with} isosurfaces, FLS \cmt{leave} the question of selecting an appropriate isovalue open. 
%
To address this issue, we use a topological approach similar to methods proposed for scalar fields~\cite{Weber2007a}. By combining FLS with scalar field topology, we introduce the concept of \emph{trait-induced merge trees}, defined as the merge tree of the trait-induced distance field in the domain.

\newtheorem*{newdef}{Definition}
\begin{newdef}
Given a multi-field $f:\Xspace\rightarrow\Aspace$ and a fixed trait $T \subset \Aspace$, let $h = h_T: \Xspace \to \Rspace$ be its trait-induced distance field. 
The \emph{trait-induced merge tree} (TIMT) is defined as the merge tree of the distance field $h$, tracking how the 
sub-level sets $h^{-1}(\infty, c]$ merge as we vary the distance parameter $c$. 
\end{newdef}

The leaves of a TIMT correspond to points in the domain that have values closest to the trait. The TIMT can be used as an interface to guide the exploration of multi-variate data, offering options to select and filter individual features.
\cmt{A crucial property of TIMT is that it is robust with respect to minor perturbations of the trait specification. Refer to Appendix:~\ref{appendix_A} for the proof of stability of TIMTs showing that the interleaving distance between TIMTs is bounded by the Hausdorff distance between the corresponding traits.}

\section{Visualization and interaction}
\label{sec:interaction}

Data sets and research questions vary, requiring different settings. To address this, we provide an interactive interface for designing traits and features. Our system supports both manual and automated trait design using Cartesian traits and learned dictionaries. For feature selection, we offer various simplification and query methods for the trait-induced merge tree, each emphasizing different aspects of the data.
The method has been implemented in an interactive visualization framework using Inviwo~\cite{Jonsson2020b} providing a large variety of rendering options. 
\begin{figure}
    \centering
        \includegraphics[width=.9\linewidth]{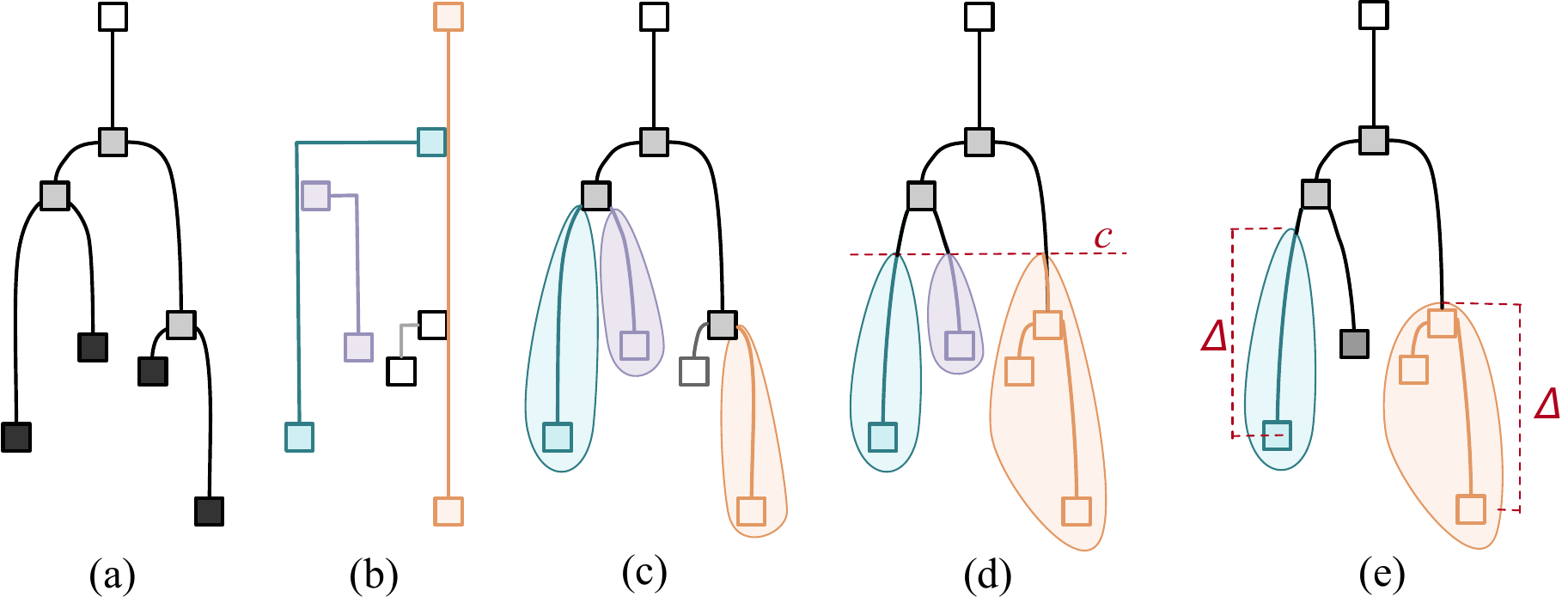}
    \caption{Query methods. (a) original tree. Users may choose between segmentation of the tree based on (b) branch decomposition, (c) leaf nodes, (d) sub-trees, and (e) crown features. 
    }
    \label{fig:query}
\end{figure}

\subsection{TIMT based features}
\label{sec:query_methods}

Based on the trait-induced merge tree we can apply different feature definitions, an overview is shown in~\fref{fig:query}. 

\paragraph{Branch decomposition} The branch decomposition is a common representation for merge trees (see \fref{fig:query}(b)). It allows for hierarchical simplification and querying. For segmentation, the user specifies a simplification threshold (persistence or hypervolume), and the method returns a domain segmentation based on this threshold. This approach always includes one branch connecting the global minimum to the maximum, which can be problematic in visualization as its vertices often enclose all others, potentially obscuring other segments and the global minimum.
\paragraph{Extremal points and their incident arcs} This method extracts the leaf nodes and their neighboring vertices and segments the domain accordingly (see \fref{fig:query}(c)). Here too, the user first specifies a simplification threshold. Unlike in the first method, it is now the merge tree itself that is simplified and then queried. This method has the advantage of highlighting every minimum separately which gives a detailed overview of the spatial distribution of minima.
\paragraph{Sub-trees} Sub-trees are extracted by first simplifying the tree as above and then cutting it at a user-specified level (see \fref{fig:query}(d)). The segments are then given by the vertices whose function value is below the threshold and which are contained within the branches that are directly affected by the cut. The result is something similar to contour forests where each segment in the domain is specified by a sub-tree originating from the cut downward.
\paragraph{Crown features} Crown features~\cite{Nilsson2022} are based on local thresholding of the merge tree. Each minimum $a$ with value $h(a)$ and a persistence value above the crown height $\Delta$ carries a crown feature. The feature is defined as the level set component at  $h(a)+\Delta$ containing the $a$.  It represents a subtree of height $\Delta$ clustering its minima. The advantage of crown features is that they prevent the main minima from overshadowing other, less prominent features, allowing these features to be extracted and visible as distinct elements.

\begin{figure*}
    \centering
    \subfloat[\label{subfig:a4_s} \small Atom-trait 4, $D_A$]{
        \includegraphics[width=0.24\textwidth]{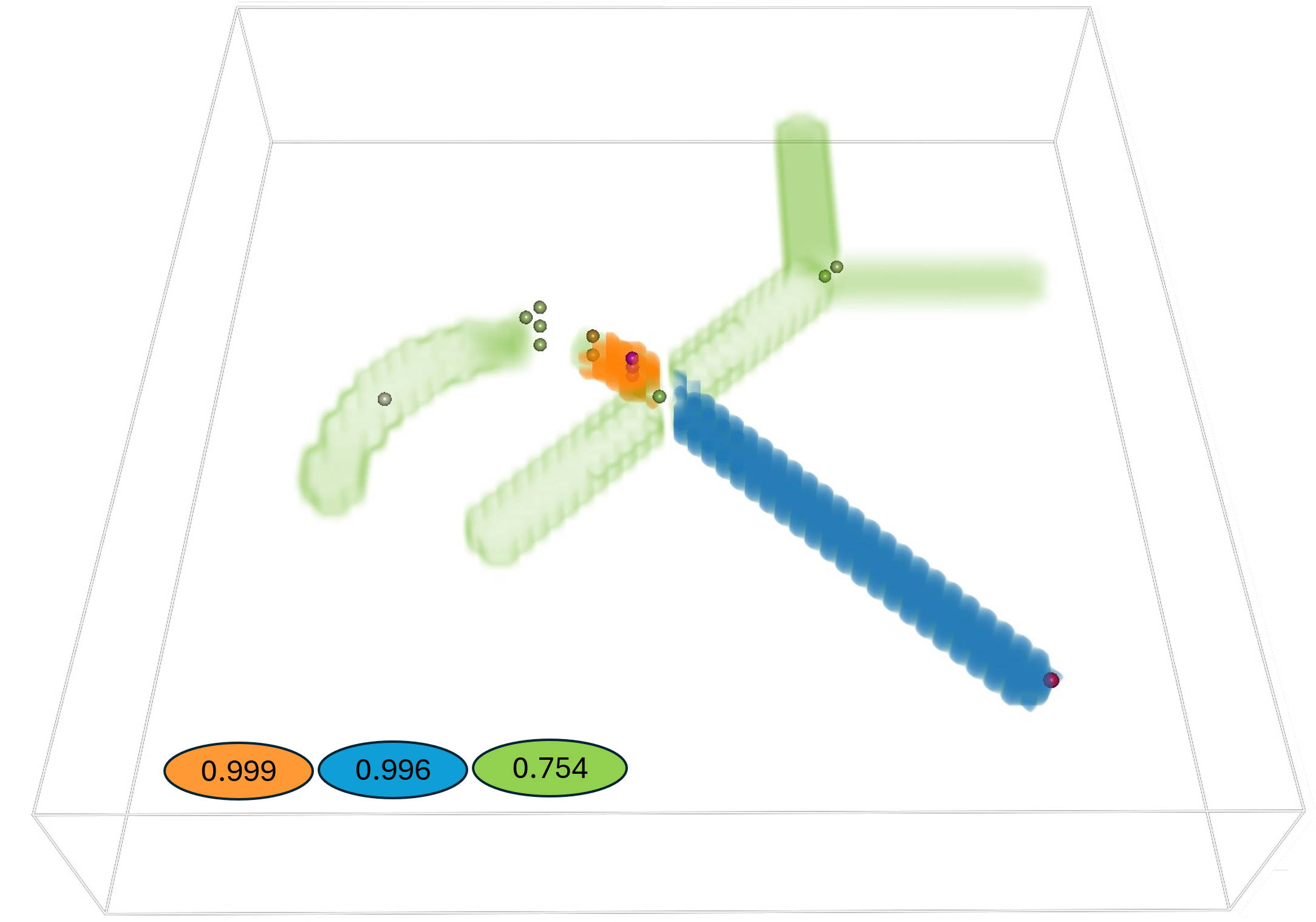}}
    \hfill
    \subfloat[\label{subfig:a6_s}\small Atom-trait 6, $D_A$]{
        \includegraphics[width=0.22\textwidth]{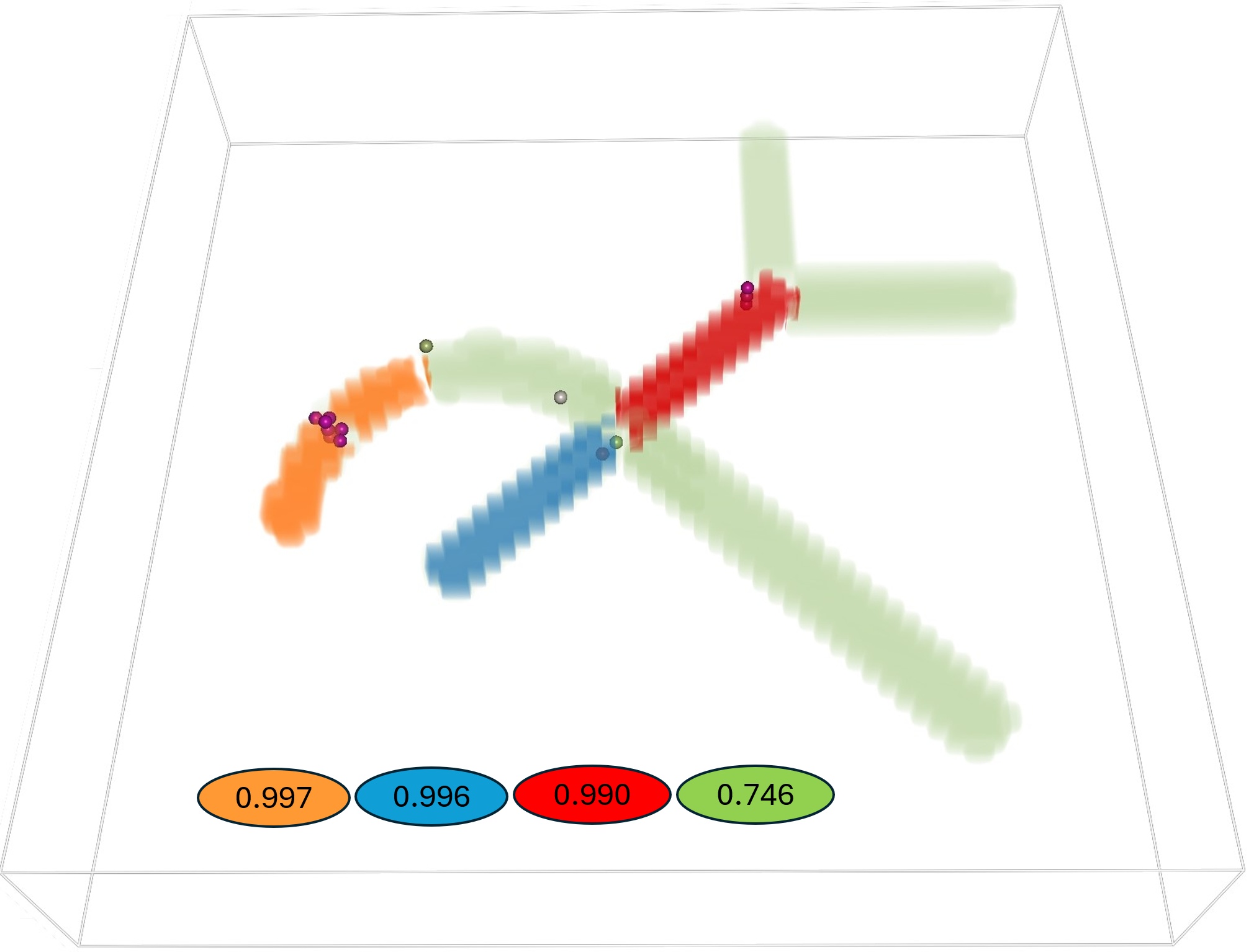}}
    \hfill
    \subfloat[\label{subfig:a1_s}\small Atom-trait 1, $D_A$]{
        \includegraphics[width=0.22\textwidth]{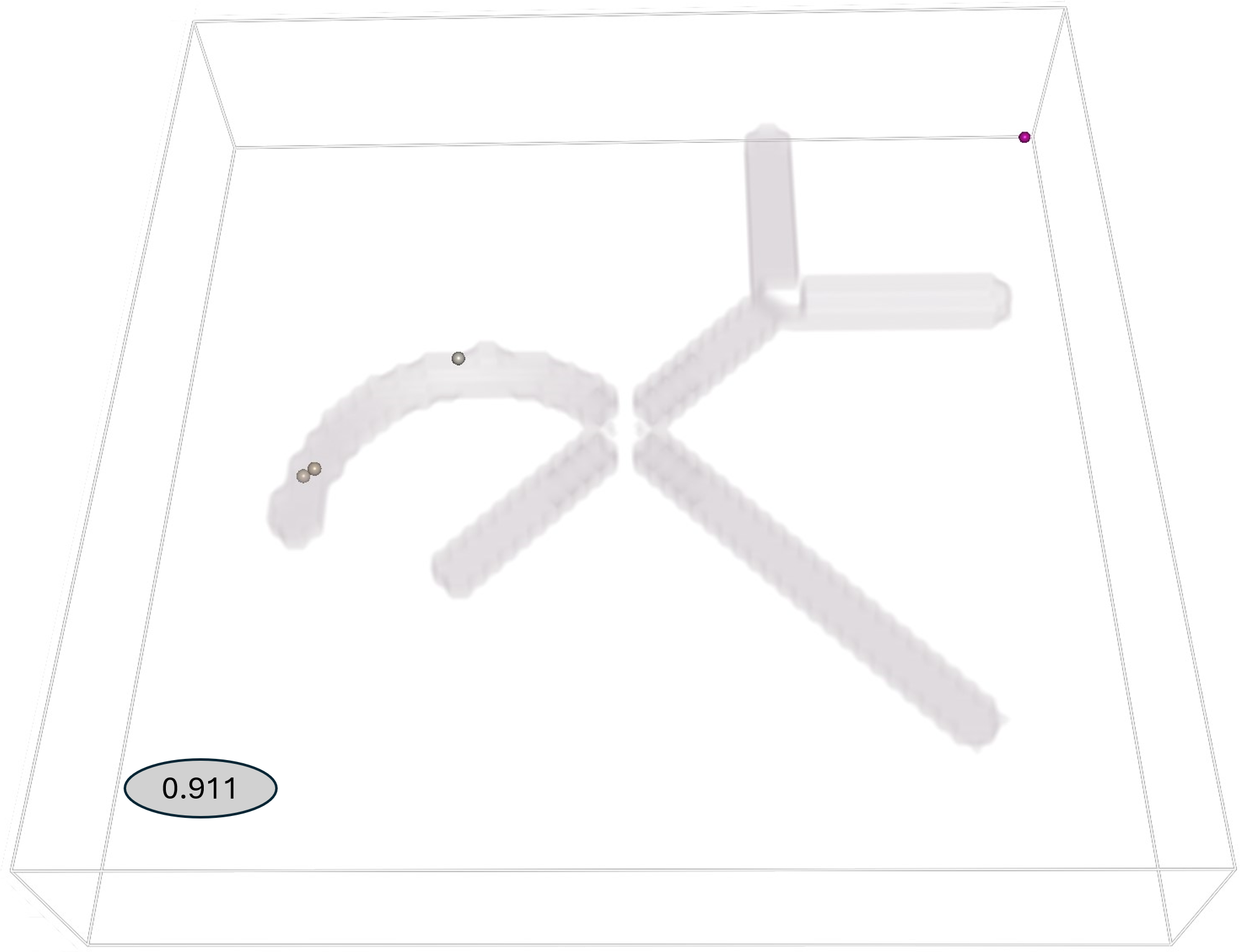}}
    \hfill
    \subfloat[\label{subfig:d30_a6_s} \small Atom-trait 15, $D_B$]{
        \includegraphics[width=0.22\textwidth]{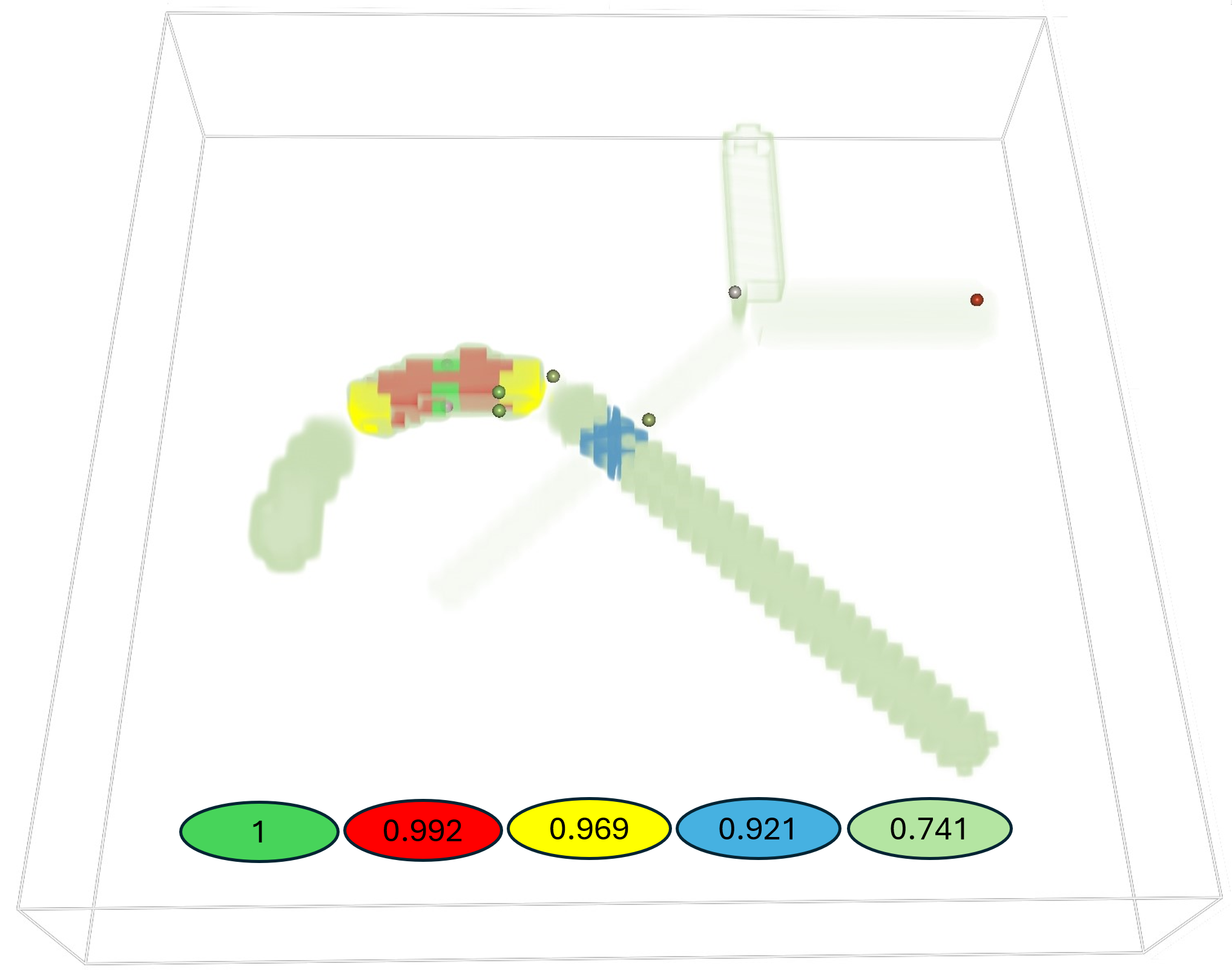}}
   \\
   
    \hspace{0.045\textwidth}        
    \caption{Using TIMTs to explore two dictionaries $D_A$ (6 atoms) and $D_B$ (30 atoms) for the phantom data set. \ct{Images (a) and (b) show the} volume renderings of selected atoms as point traits, highlighting segments with the highest similarity in color. Light green parts with low similarity values are rendered to provide context. (c) Atom 1 mainly represents the background, which is least similar to the trait shown. Image (d) demonstrates that the selected atom cannot be directly associated with one direction.}
    \label{fig:phantom_segmentation}
\end{figure*}


\subsection{Visualization and segmented volume rendering}
\cmt{We provide the option to visualize the extracted features by coloring the voxels according to their segmentation,} as shown in~\fref{fig:mol-symmetric}.
Alternatively, a direct volume rendering (DVR) of the segmented trait-induced distance field is possible. Here, random colors are automatically assigned to all segments, which can be toggled on and off.  
Users can use a global transfer function or specify transfer functions separately for individual segments. Depending on the selected feature, different isovalues are recommended to highlight the respective features.

\subsection{Interaction}
\cmt{We provide two ways of interacting with the segmentation of the domain for locating the feature.} 
The first interface consists of a legend positioned below the 3D rendering. The legend contains a button for every segment in the data. Clicking on such a button will toggle the voxel-wise rendering of that segment. 
Active buttons are highlighted with a red boundary. For navigation, the colors of the buttons and the rendered segments correspond to each other. Additionally, the buttons show the value of the segment's minimum, which measures the distance to the trait. This way, the user can select segments based on their distance to the trait.
The second interface is a volume slice, see \fref{fig:tpl}. After positioning the slice in the volume, the user can select segments on the slice by clicking on them. After selecting segments, the user may extract a connected surface of all selected segments either for further analysis or for rendering it in a DVR context.

\section{Case studies}
\label{sec:results}

We demonstrate the powerful utility and generality of our proposed framework through use cases across various applications, including scalar, vector, and tensor field datasets.

\subsection{Understanding of dictionary learning using TIMT}
\label{cs:A}
In Section~\ref{sec:point-trait-sparse}, we proposed using sparse dictionaries to design point traits. In this initial case study, we demonstrate that TIMTs can also provide a deeper understanding of the learned dictionary. By integrating dictionary learning with TIMT, we explore how atoms capture and represent the underlying structure and features of the data, enhancing the interpretability of the learned dictionary.
%
\begin{figure}[!t]
\centering

   \subfloat[\small Atom glyphs showing the 60 measurement directions]{
        \includegraphics[width=0.45\textwidth]{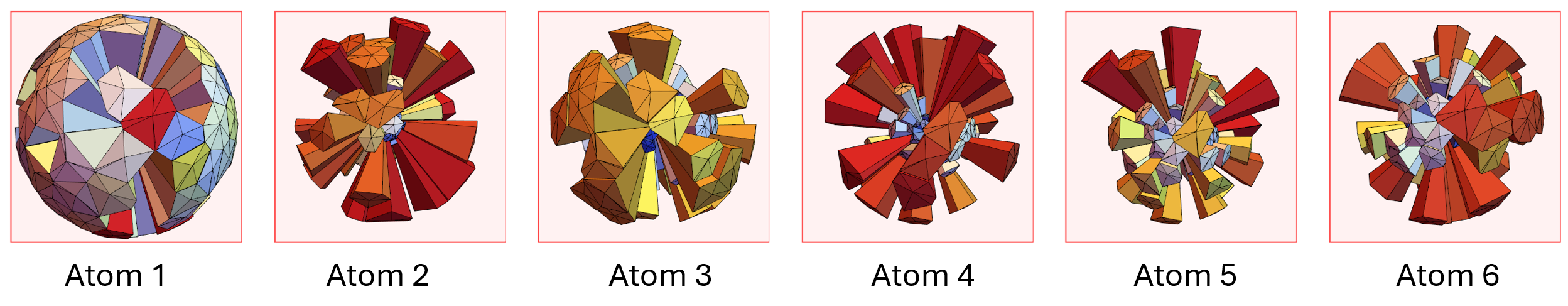}}
        \hfill    
 \subfloat[\label{subfig:atom_odf}\small Atoms as ODF (Orientation Distribution Function) ]{
        \includegraphics[width=0.45\textwidth ]{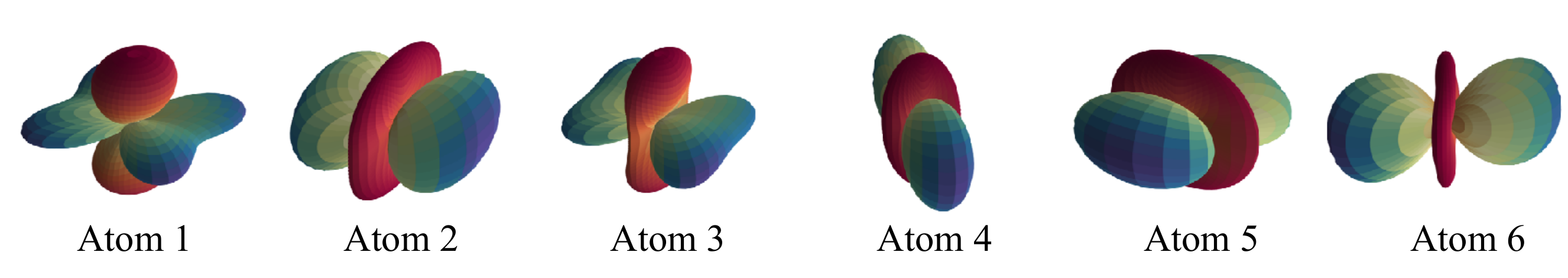}}
\caption{A compact dictionary of 6 atoms with a sparsity level of 3 is used to explore the diffusion phantom data. \cmt{These atoms reveal the essential patterns extracted from the data and each atom captures distinct fiber features (fiber orientations, fiber crossing).} 
In (a), the 60 diffusion directions of the learned values are visualized as a glyph, where red and blue represent higher and lower values, respectively. In (b), the orientation distribution function (ODF) is depicted using spherical harmonics and the Funk-Radon transform~\cite{descoteaux2007regularized}, commonly used to indicate fiber bundle orientation, with red and blue denoting lower and higher values, respectively.}
\label{dictionary_phantom}
\end{figure}

%
We utilize the phantom HARDI dataset~\cite{phantom}, a simple yet high-dimensional dataset sized $71 \times 71 \times 15 \times 60$, designed to simulate diffusion behaviors of fiber bundles within the brain, as described in~\cite{leemans2005mathematical}. This dataset represents a fiber structure that features cross-sections, splittings, and curved regions, serving as a benchmark for testing fiber tractography. Each voxel in this dataset contains 60 diffusion-weighted measurements. According to the simple structure represented by the data, it is naturally sparse.
The goal of this case study is to explore the properties of the learned dictionary by comparing different parameter settings. 

\ct{
Determining the optimal number of atoms for a dictionary lacks a universal answer. While guidelines exist for compression-focused tasks, they do not apply when prioritizing the representational power of atoms, where the goal is to 
find a dictionary to effectively represent this phantom data.
This challenge resembles finding the optimal number of clusters in clustering.
To address this, we tested and compared dictionaries of varying sizes. Therefore, we treat atoms as point traits (atom-traits) to construct FLS, highlighting areas in the volume that exhibited high similarity with the original atoms. Starting with a dictionary consisting of 30 atoms ($D_{B}$), we quickly observed that there is high redundancy between the FLS for some of the atoms. To get a better understanding of the dictionary, we computed pairwise similarities of the atoms, which confirmed our observation. These similarity matrices show that the atoms form approximately six clusters (details, including the similarity matrices, can be found in the Appendix:~\ref{appendix_B}). Selecting one representative atom-trait from each cluster provided good data coverage, capturing distinct features—specifically, fiber directions. However, results for two atoms from the same cluster were very similar. A reduced dictionary ($D_{C}$) of 10 atoms produced similar results, leading us to finalize a six-atom dictionary ($D_{A}$) for further analysis.}The atoms for this configuration are illustrated using spherical harmonics in~\fref{subfig:atom_odf}.

\ct{Some of the} results for the different dictionaries are shown in ~\fref{fig:phantom_segmentation}, one with six atoms, denoted as $D_A$, and another with 30 atoms, denoted as $D_B$. (\ct{More results can be found in the Appendix:~\ref{appendix_B}.}) In $D_A$, four atoms distinctly capture different directions. \fref{subfig:a4_s} illustrates Atom-trait 4, which represents a distinct direction with two connected components (orange and blue). 
In contrast,~\fref{subfig:a6_s} shows Atom-tait 6 guiding a different direction with three connected components. \fref{subfig:a1_s} effectively filters out the background to capture the primary structure of the data. In $D_B$ we only show results for one atom, ~\fref{subfig:d30_a6_s} mainly highlights the curved region of the fiber with a few clusters and one cluster with a lower similarity value in the region of the crossing fibers. \ct{This demonstrates that a dictionary with too many atoms can lead to overfitting,}  resulting in less expressive atoms.

These results show that designing a tailored dictionary specific to a dataset is crucial for achieving an efficient and compact representation, \ct{which is} foundational for using the dictionary as a trait to explore features. The characteristics, dimensionality, underlying structure, and sparseness of the data significantly impact the dictionary's performance.

\begin{figure}[ht]
    \centering
     \includegraphics[width=0.9\linewidth]{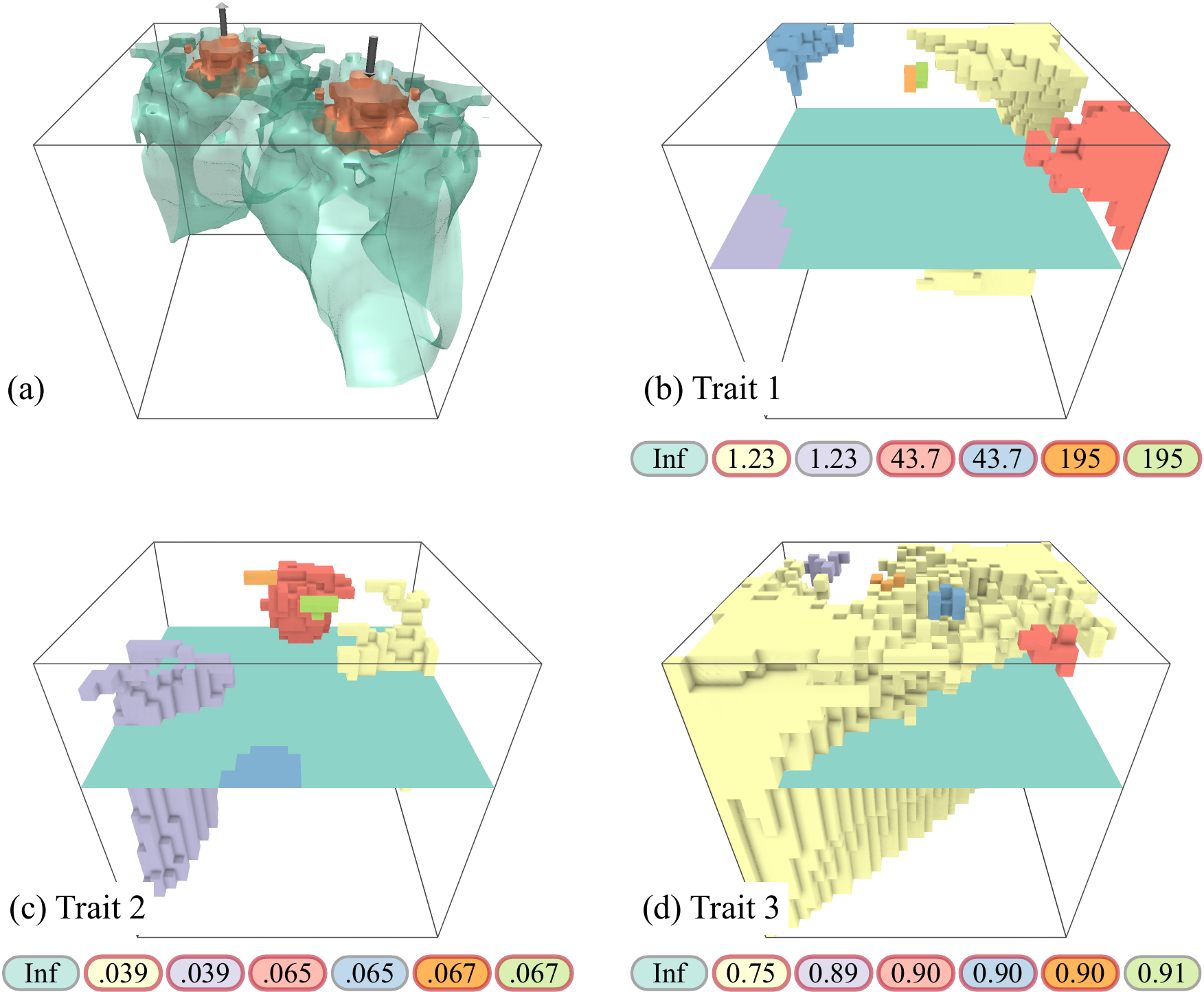}
    \caption{
        (a) Illustration of the two-point-load data set. The arrows indicate the forces applied to the block of metal. The isosurfaces display the anisotropy that occurs in the material.
        (b) Trait 1: eigenvalues equal to zero.
        (c) Trait 2: high isotropy.
        (d) Trait 3: planar anisotropy and one high principal eigenvalue.
        Shown is the segmentation of the distance field. The active segment buttons are highlighted with a red boundary. \ct{Note that the visibility of the purple volume in (b) is turned off.}}
    \label{fig:tpl}
\end{figure}

\begin{figure*}
    \centering
    \includegraphics[width=0.95\linewidth]{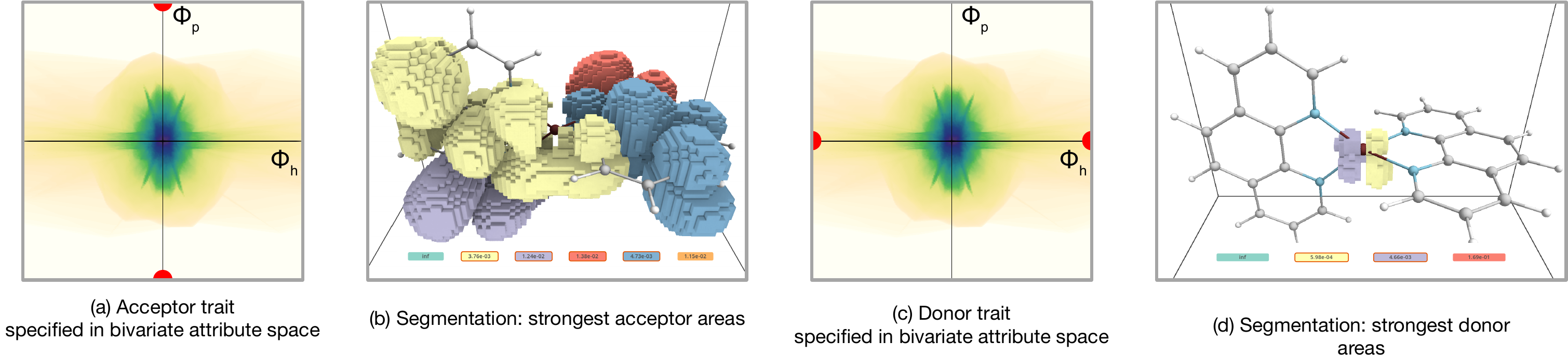}
    \caption{Molecular electronic transition data set (copper complex) with symmetric ligands. Images (a) and (c) illustrate the CSP plots together with the \cmt{acceptor and donor} traits, respectively (red dots).
    Images (b) and (d) show \cmt{acceptor} regions respective \cmt{donor} regions.
        The merge tree has been simplified using the hypervolume metric. The regions correspond to the $n$ lowest leaves in the tree.}
    \label{fig:mol-symmetric}
%
    \centering
    \includegraphics[width=0.95\linewidth]{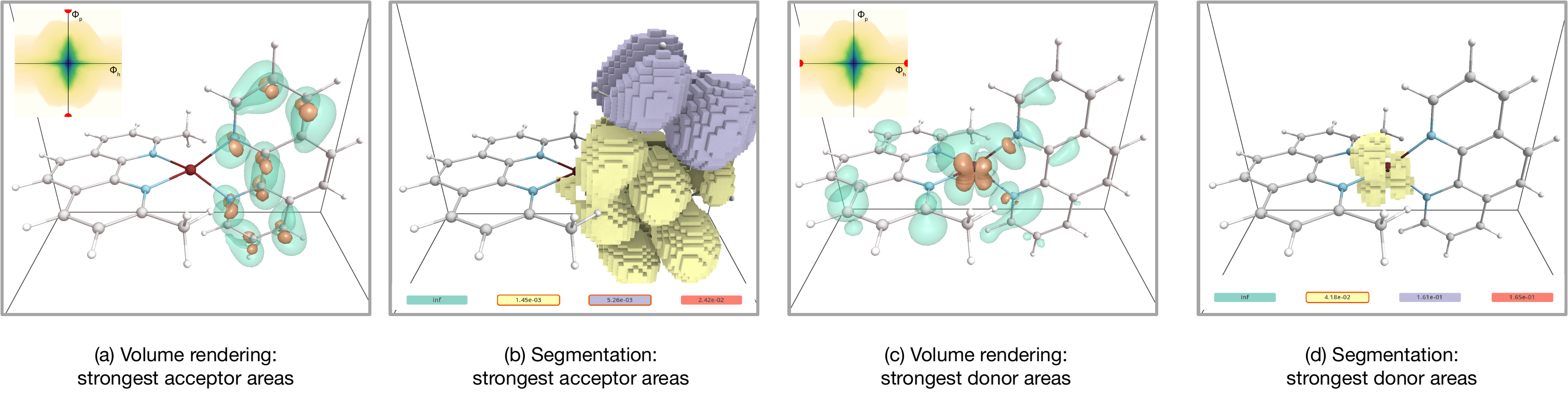}
    \caption{Molecular electronic transition data set (copper complex) with asymmetric ligands. The CSP plots and the \cmt{acceptor and donor} traits are shown as inlays in images (a) and (c), respectively (red dots).
      Images (a) and (c) exhibit the volume rendering of the \cmt{acceptor and donor} distance field. 
      Images (b) and (d) show the segmentation for the \cmt{acceptor} regions and \cmt{donor} regions directly. The merge tree has been simplified using the hypervolume metric. The regions correspond to the $n$ lowest leaves in the tree.}
    \label{fig:mol-asymmetric}
\end{figure*}


\subsection{Tensor field analysis - Two-point load}
\label{sec:tensor}

Since FLS was originally developed for tensor fields, our \cmt{second} example uses them. Using an individual tensor as a trait, FLS introduces \emph{tensor level-sets} as the zero level-set of a point trait representing one tensor. Combined with trait-induced merge trees, this creates a new concept of \emph{tensor field topology}.
To verify the method, the first case study examines a numerical material simulation of stresses in a solid block. Two forces are applied to the top, one pulling and one pushing, as shown in \fref{fig:tpl}(a). This data set is referred to as \emph{two-point-load}.
The simulation output is a stress tensor field with symmetric tensors having six degrees of freedom. The expected stress has a high linear \cmt{anisotropy} at impact points, planar anisotropy along the midsection between these points, and low stresses at regions far away from the impact points.

An appropriate attribute space for related features is spanned by the three eigenvectors, referred to in material sciences as principal stresses $\lambda_i$, $i=1,2,3$, ordered as $\lambda_1\ge\lambda_2\ge\lambda_3$, or anisotropy measures. Anisotropy measures of interest are linear $c_l$, planar $c_p$, and spherical $c_s$ anisotropy given by $c_l=\frac{\lambda_1-\lambda_2}{\lambda}$, $c_p=\frac{2(\lambda_2-\lambda_3)}{\lambda}$, and $c_s=\frac{3\lambda_3}{\lambda}$, respectively, with $\lambda=\lambda_1+\lambda_2+\lambda_3$~\cite{Westin1997}. A common measure in material sciences is the maximum shear stress, defined as $\lambda_1-\lambda_3$. \fref{fig:tpl}(a) shows two level-sets of this measure.
In our example, the traits were defined to match these criteria to verify the resulting regions.
For the first trait, all three principal stresses were set to zero. \fref{fig:tpl}(b) shows regions close to this behavior, aligning with our expectations since these regions are farthest from the impact points.
\fref{fig:tpl}(c) shows regions with isotropic behavior, where the trait was set to high spherical anisotropy ($c_s$) and low linear and planar anisotropies ($c_l$ and $c_p$). Compared to \fref{fig:tpl}(b), this trait highlights neighboring regions. This makes sense, as areas unaffected by the forces have a small band of isotropic behavior around them before a more distinctive stress distribution emerges.
Lastly, the third trait was set to highly planar behavior ($c_p$) coupled with a high major principal stress value ($\lambda_1$). \fref{fig:tpl}(d) depicts the resulting regions. As expected, the middle section corresponds to this trait.
%

%


\subsection{Acceptor-donor regions in molecular electronic transitions}
\label{sec:chemistry}
In this case study, we analyze molecular electronic transitions using ~feature level sets and trait-induced merge trees. The electronic structure of a molecule changes its interaction with light, represented by two scalar fields, $\Phi_h$ and $\Phi_p$, denoting the spatial distribution of electrons before and after photon absorption during the transition~\cite{Masood2021Transitions}. 
Chemists are interested in how the localization of electronic distribution changes during the transition and how different molecular configurations affect these transitions. It is crucial to identify which parts of the molecule lose and gain charge, acting as donor and acceptor regions, respectively. 
Recently, Sharma et al.~\cite{sharma2021segmentation, sharma2023continuous} proposed treating these scalar fields as a single multi-field and applying bi-variate analysis. They suggested that examining patterns in the continuous scatter plots~\cite{Bachthaler2008CSP} of the bi-variate field for the entire molecule or its sub-regions can reveal donor and acceptor behavior.

Here, we examine electronic transitions in two copper complexes with slightly different configurations: one with symmetric ligands (identical molecular groups around the copper atom) and one with asymmetric ligands (different molecular groups). The goal is to identify and compare the donor and acceptor regions in these complexes.
A donor region has a higher concentration of electronic density before the transition compared to after, characterized by the condition $|\Phi_h| > |\Phi_p|$. In the case of ideal donor behavior, the donor trait can be defined by points $(\max|\Phi_h|, 0)$ and $(-\max|\Phi_h|, 0)$ in the bi-variate space of $\Phi_h \times \Phi_p$. Similarly, the acceptor trait is defined by points $(0, \max|\Phi_p|)$ and $(0, -\max|\Phi_p|)$. These traits are indicated by red disks in the continuous scatter plots (CSPs) shown in \fref{fig:mol-symmetric} and \fref{fig:mol-asymmetric}.

Using these point traits for donor and acceptor regions, we extract feature level sets for both molecules. In the case of symmetric ligands, strong donor behavior is expected around the central copper atom, as copper is known to be a strong donor. The acceptor regions are expected to be equally distributed over the two identical surrounding molecular groups, as there is no preference for one group over the other. We applied TIMTs to analyze this behavior by querying the regions corresponding to the leaf nodes with the lowest values in the donor TIMT. This allowed us to automatically identify the region around the copper atom \fref{fig:mol-symmetric}(d).  Similarly, for the acceptor trait, the regions were distributed over the two groups around the copper atom \fref{fig:mol-symmetric}(b). Interestingly, the regions corresponding to leaves with the lowest values captured the two molecular groups as separate regions in the TIMT segmentation. This automatic subdivision of the acceptor region into sub-regions matching the chemical subgroups was much appreciated by our collaborator.

Next, we analyze the second copper complex with asymmetric ligands. As before, the donor region is concentrated on the central copper atom, as can be seen in the volume rendering and segmentation results in \fref{fig:mol-asymmetric}(c) and (d). However, the acceptor region now concentrates on only one of the surrounding molecular groups \fref{fig:mol-asymmetric}(a) and (b). This indicates a preference for electronic charge transfer to one group over the other in the case of asymmetric ligands, a behavior of particular interest to chemists. Lastly, we want to point out that the continuous scatter plots for both the complexes look quite similar, making it difficult to distinguish between the two transitions based on these plots alone, as seen in \fref{fig:mol-symmetric}(a) and in the inlays of \fref{fig:mol-asymmetric}(a). This demonstrates the importance of exploring transition behavior in the spatial domain, which is facilitated by the feature level sets and domain-specific trait-induced merge tree segmentations.


\subsection{Vortex re-connection}
\label{sec:flow}

\begin{figure}
    \centering
       \includegraphics[width=0.8\linewidth]{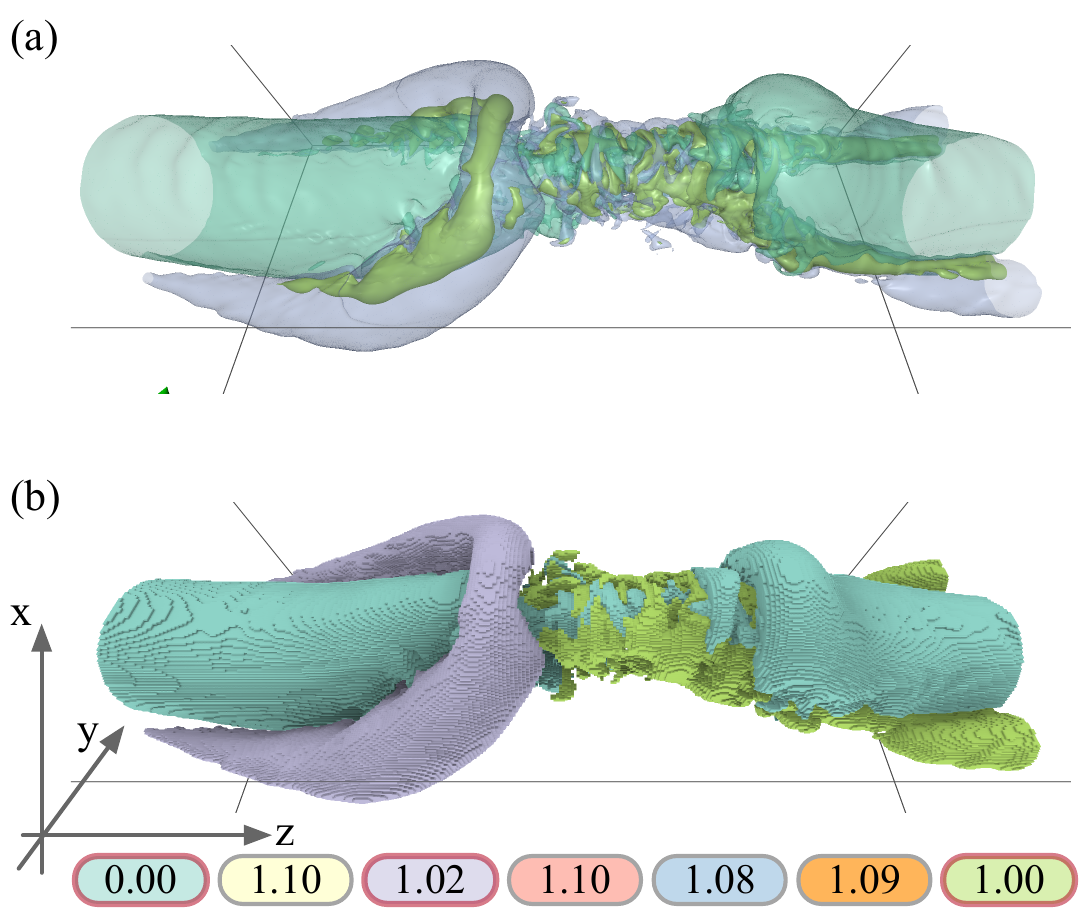}
    \caption{Vortex re-connection simulation.
       (a) Volume rendering of the distance field for a manually designed trait.
        (b) The corresponding segmentation using branch decomposition is shown. The purple segment highlights the horseshoe.}
    \label{fig:flow}
\end{figure}

Vortices are rotational fluid flow regions around a core line defined by a curve in space, playing a crucial role in fluid dynamics. Various methods exist for identifying and extracting vortex structures and their core lines~\cite{Gunther2018Vortex, Kasten2011Acceleration}. However, understanding the dynamics of vortices—particularly how they evolve and interact—is an ongoing area of research~\cite{Bujack2020FlowSTAR, Kasten2012VortexMerge}. In this case study, we focus on data involving two parallel counter-rotating vortices that interact over time, leading to a re-connection event. This phenomenon is commonly observed in vortices shed by aircraft wing tips, sometimes visible in the sky as condensation trails. Initially parallel, these trails undergo a \emph{re-connection} event, forming closed loop-like structures before dissipating. Our collaborators conducted numerical simulations to study such re-connection events.  One of their main objectives is to understand the formation of \emph{horseshoe-like} structures during the re-connection, which grow over time and eventually lead to the formation of disconnected loops. Identifying and analyzing this horseshoe structure is a key task in the analysis of this simulation data.

This task is challenging because there is no established definition for automatically extracting horseshoe structures. Therefore, they are ideal candidates for using FLSs and TIMTs to explore different trait and feature definitions. Based on input from our collaborators, we selected a time step after the re-connection event where the horseshoe structure is visible.
One characteristic of the horseshoe is its appearance as a weaker vortex in a plane orthogonal to the two parallel vortices. As vortices in general, they are characterized by a lower pressure at their cores. We utilize these observations to design a trait that can capture the horseshoe structure. Given that the two parallel vortices lie along the Z-axis and are separated in the $Y$ direction, we expect the horseshoe to reside in the $XY$ plane, with its core approximately aligned with the $Y$-axis.

We start with manually formalizing these properties as a trait by considering the velocity vector field $\textbf{v}$, composed of the three velocity components $(v_x, v_y, v_z)$ and the pressure field as a \cmt{multi-field}. The trait is defined by high absolute values of $v_x$ and $v_z$, combined with low absolute values of $v_y$ and low-pressure values.~\fref{fig:flow} shows the results obtained for this trait. 
The purple segment captures the horseshoe structure that is closest to our trait. For the segmentation, we employ the branch decomposition segmentation derived from the trait-induced merge tree. 
This result indicates that feature level sets and trait-induced merge trees can be employed to extract complex features from multi-fields using simple queries, which would otherwise be challenging to identify.

\begin{figure}[t]
\centering
    \includegraphics[width=0.9\linewidth]{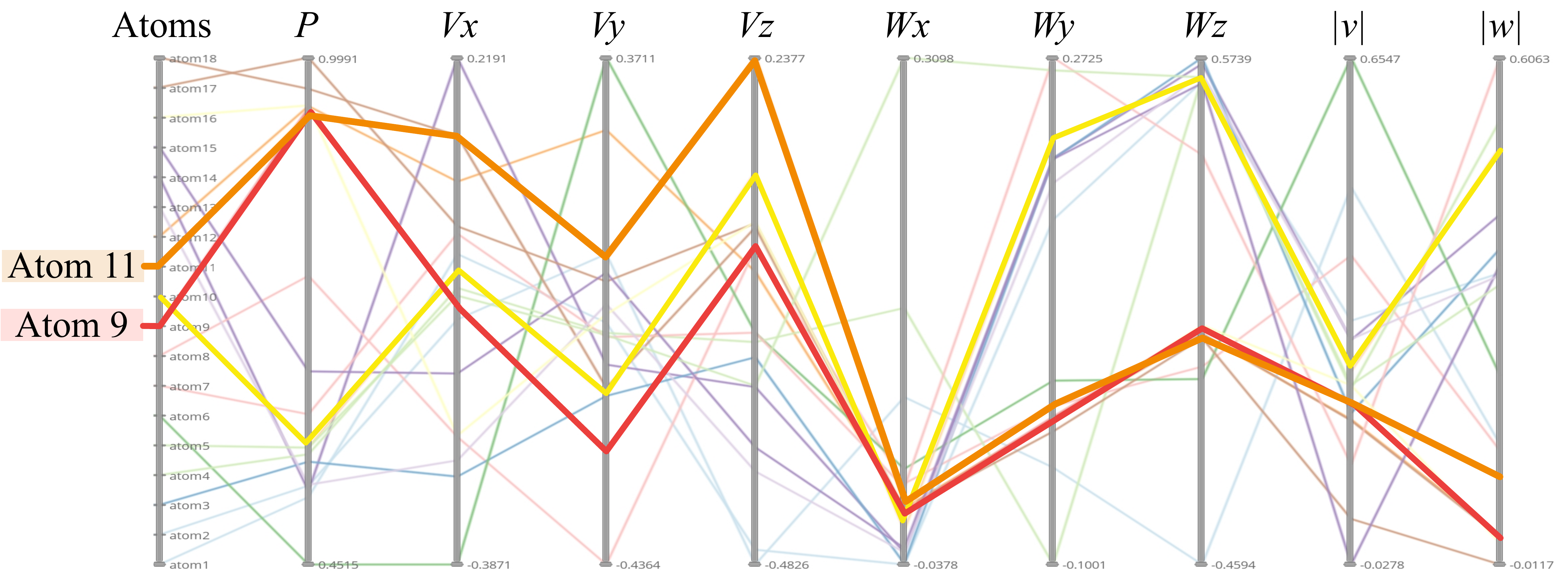} %
    \caption{Illustration of the dictionary of 18 atoms using parallel coordinates, highlighting their attributes.
\cmt{Each column represents the values for the different attributes of the atoms. Negative values in $|v|$ and $|w|$ arise from the optimization process used to construct the atoms. These can be avoided by incorporating the positivity constraints into the optimization process.} Atoms 9 and 11 are specifically highlighted. Atom 11 represents a horseshoe structure, characterized by a correlation with a high value in the $v_z$ direction, low vorticity in the $x$ direction, and mid-range vorticity in the $z$ direction.
}%
    \label{fig:para_coor}%
\end{figure}

\begin{figure}[b]
    \centering
    \subfloat[\label{subfig:a11_horseshoe} ]{{\includegraphics[width=0.7\linewidth]{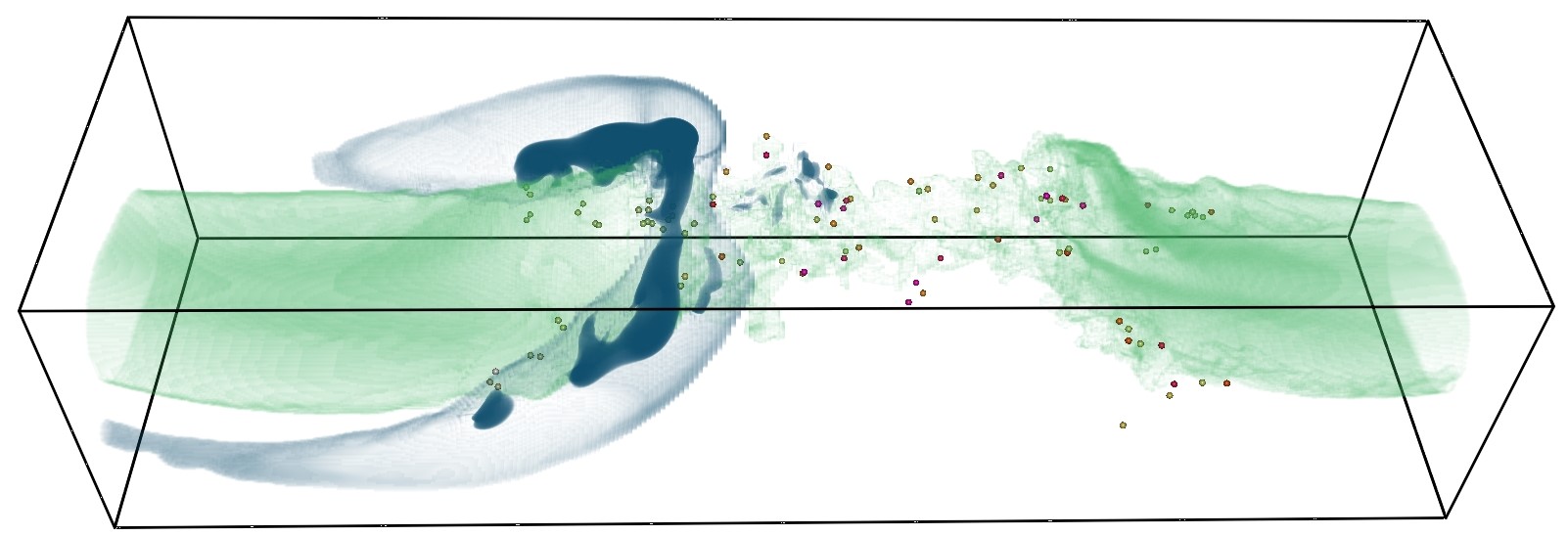} }}%
    \hfill
    \subfloat[\label{subfig:vol_render} ]{{\includegraphics[width=0.7\linewidth]{
    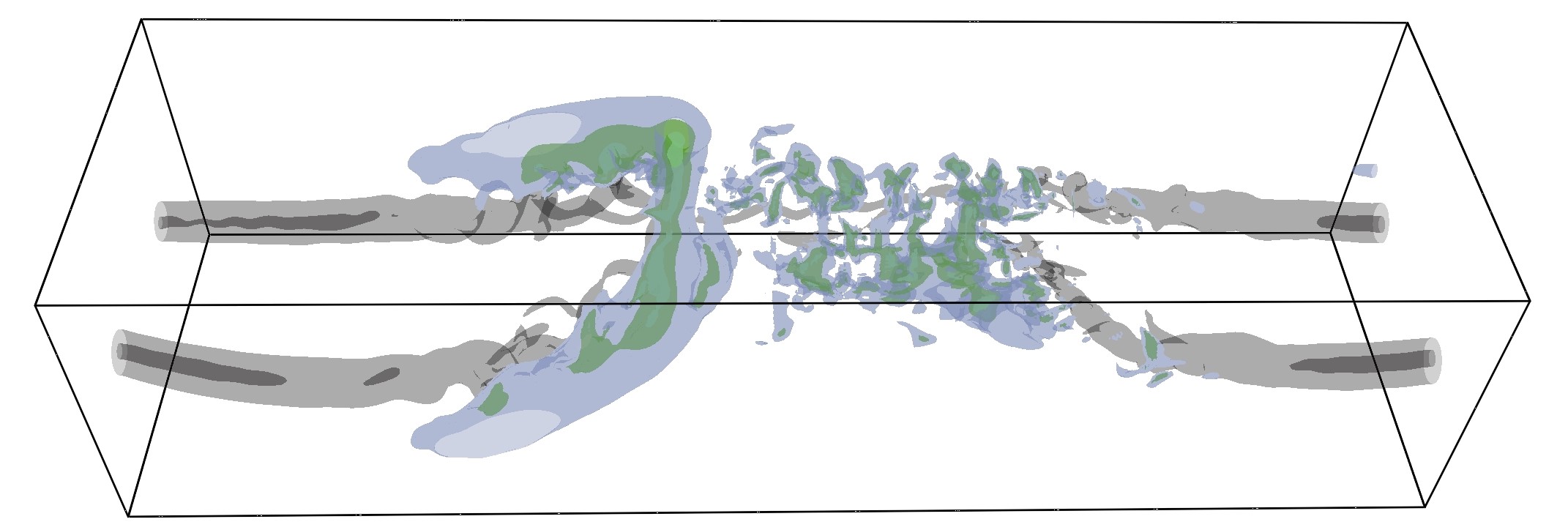
    } }}%
    \caption{Example of horseshoe structure that associated with trait 11. Image \protect\subref{subfig:a11_horseshoe} demonstrates the segmentations using atom 11 as a trait for TIMT. Image \protect\subref{subfig:vol_render} shows the horseshoe structure automatically extracted by dictionary learning with original parallel vortices rendered in a different color set as a reference.}%
    \label{fig:horseshoe}%
\end{figure}

\begin{figure}[ht]
    \centering
    {\includegraphics[width=0.7\linewidth]{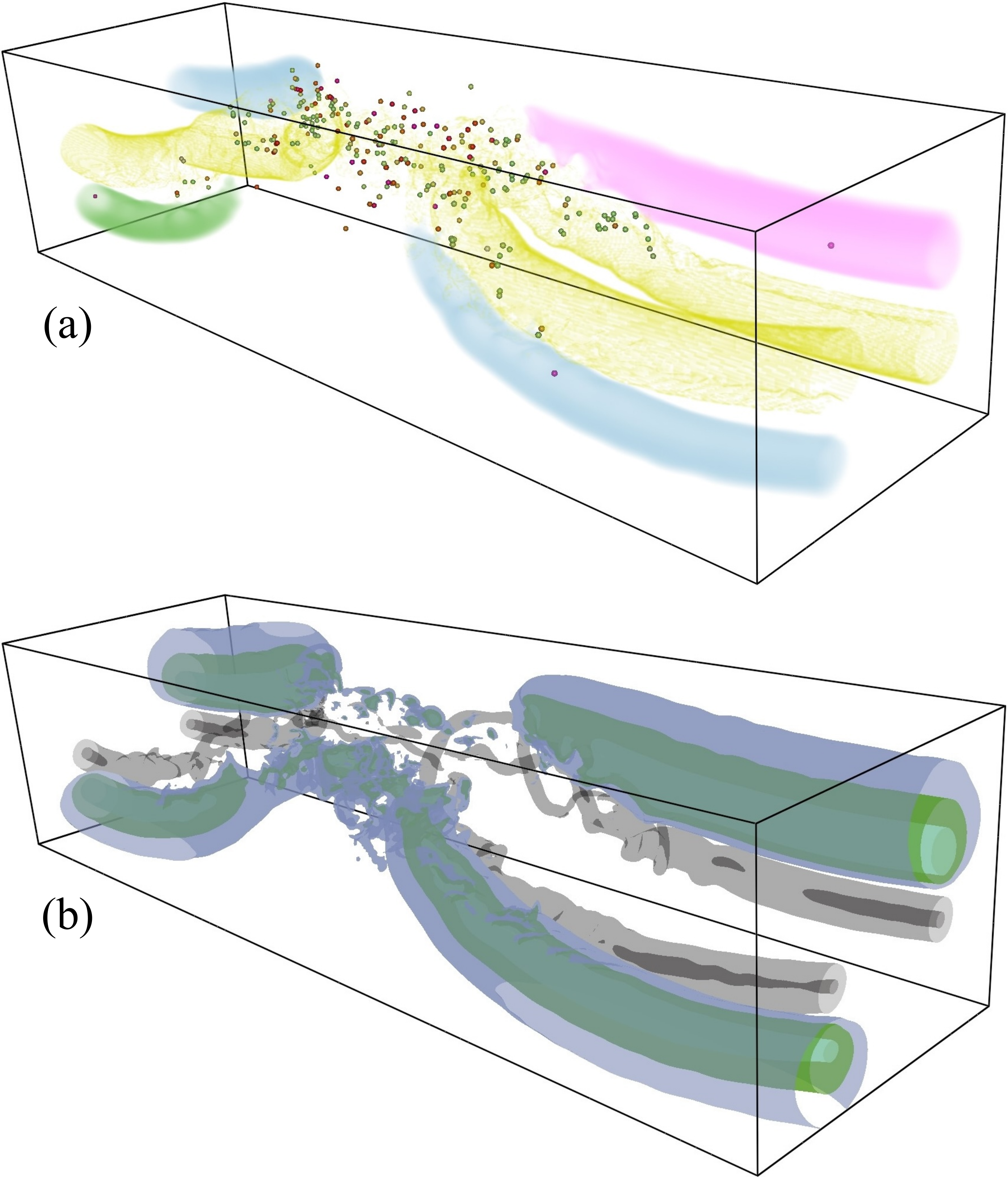} }
    \caption{Example of vortex rotation direction component associated with trait 9. Image (a) exhibits segmentations of the largest components of computing TIMT. Additionally, image (b) presents the original vortex flow in grey and black and the automatically extracted vortex rotation direction component in purple and green by dictionary training.
    }
    \label{fig:vortex-rotation}
\end{figure}

In a second experiment, we investigate \emph{atom-traits} from a learned dictionary, enabling the simultaneous visualization of multiple features and providing a more comprehensive understanding of the data. Here, we configure our multi-field from pressure, the three velocity components $(v_x, v_y, v_z)$, three vorticity components  $(\omega_x, \omega_y, \omega_z)$, and their magnitudes. Given that this dataset contains both scalar and vector fields, each with distinct value ranges and units, we utilized individually scaled or normalized values for each field. We trained the dictionary with the recommended number of atoms, twice the number of dimensions, which is 18. All atoms' parameters are displayed in the parallel coordinates plot~\fref{fig:para_coor}.
By examining \emph{atom-traits}, we identified atom 11 that captures the horseshoe structure similarly to the manually designed trait. 
 \fref{subfig:a11_horseshoe} shows a rendering of regions closest to this trait 11, such as zones of a horseshoe and the central area of the data set, similar to the manual trait~\fref{fig:flow}.
 \fref{subfig:vol_render} renders the main segment of atom 11 together with regions of high vorticity emphasizing the parallel vortices as a reference.

Looking at other atoms reveals that they cluster certain directional behaviors of the velocity or vorticity, highlighting areas along the main vortices but not their centers. An example is shown in~\fref{fig:vortex-rotation}(a), illustrating the four main clusters for atom-trait 9. \fref{fig:vortex-rotation}(b) shows the vortex core in gray relative to the highest values of the similarity field.


These examples demonstrate that integrating dictionary learning with TIMT for the analysis of complex flow structures can reveal interesting patterns within the data. However, it also shows that not every atom has a strong physical meaning and requires further investigation.

\begin{figure*}[ht]
    \centering
    \subfloat[\small Cloud-triggering features]{%
        \includegraphics[width=0.32\textwidth]{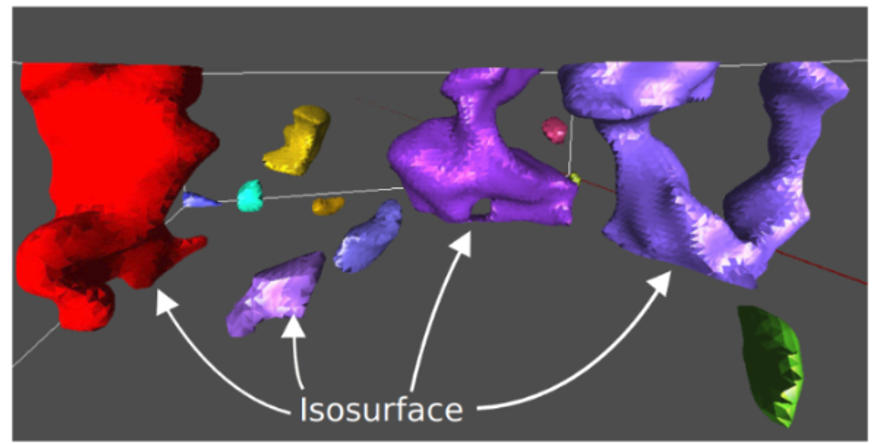}
        \label{fig:subfig_1}%
    }
    \hfill
    \subfloat[\small Cartesian Fiber Surface]{%
        \includegraphics[width=0.32\textwidth]{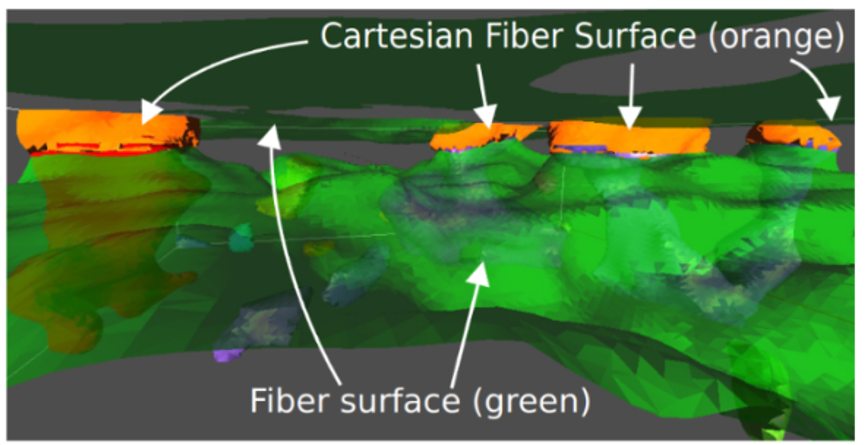}%
        \label{fig:subfig_2}%
    }
    \hfill
    \subfloat[\small Continuous Scatterplot]{%
        \includegraphics[width=0.25\textwidth]{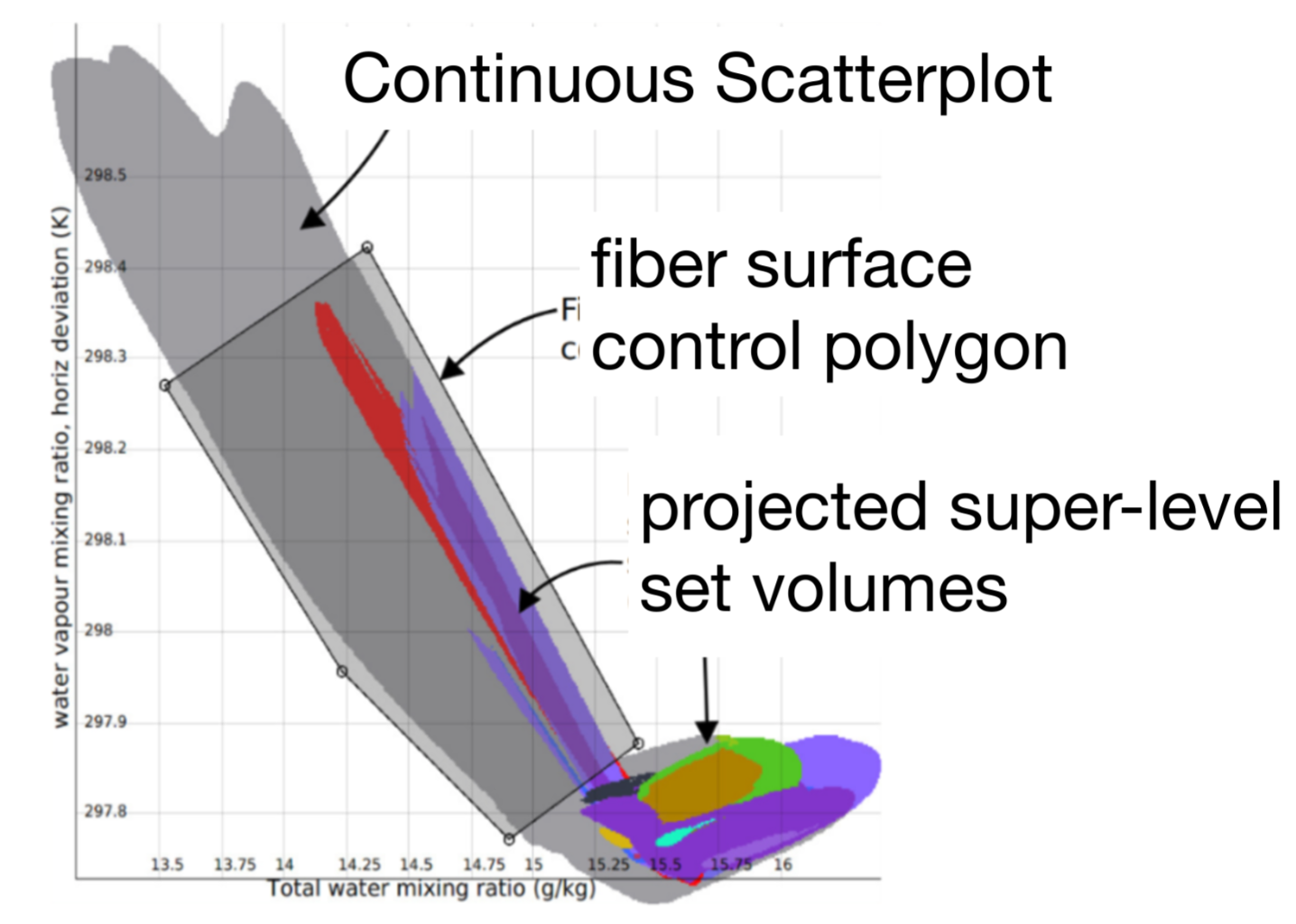}%
        \label{fig:subfig_3}%
    }
    \caption{
        Convective cloud formation data set. Image (a) illustrates an isosurface of the cloud-triggering structures.  Image (c) shows the continuous scatterplot of temperature and humidity, where the volumes of the structures are projected and overlaid. Fiber surfaces in (b) can then be defined via a control polygon and compared to the isosurface to examine the properties of parts of the objects. Image (c) visualizes the parts of the features defined by the isosurfaces with properties inside the control polygon using a Cartesian fiber surface. This shows that the parts of the cloud-triggering features with lower humidity and higher temperature are at the top of the spatial domain.
    }
    \label{fig:cartesian-fs}
\end{figure*}

\begin{figure*}[ht]
    \centering
    \subfloat[\small Polygon Trait]{%
        \includegraphics[width=0.3\textwidth]{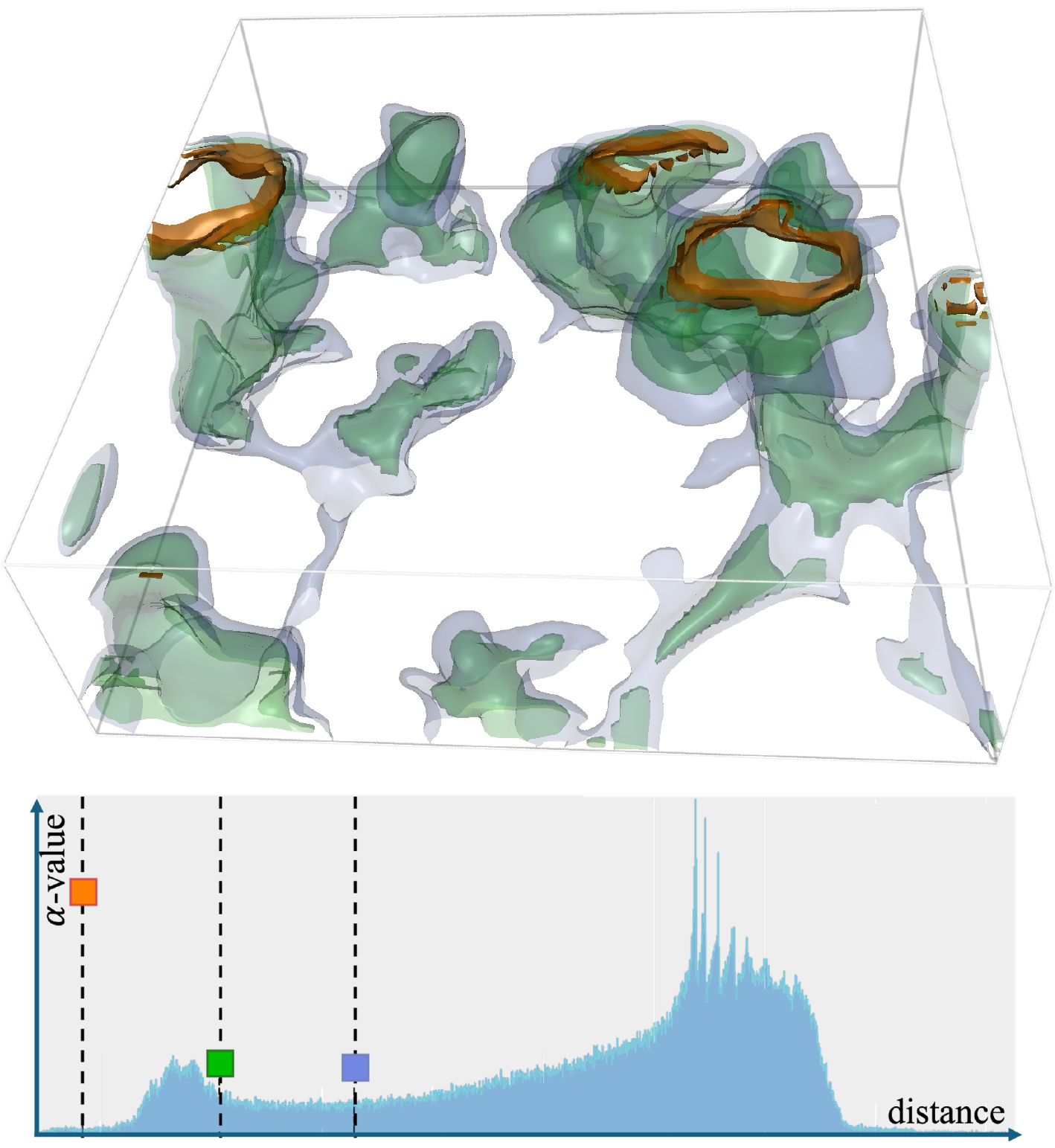}%
        \label{fig:subfig1}%
    }
    \hfill
    \subfloat[\small Point Trait]{%
        \includegraphics[width=0.3\textwidth]{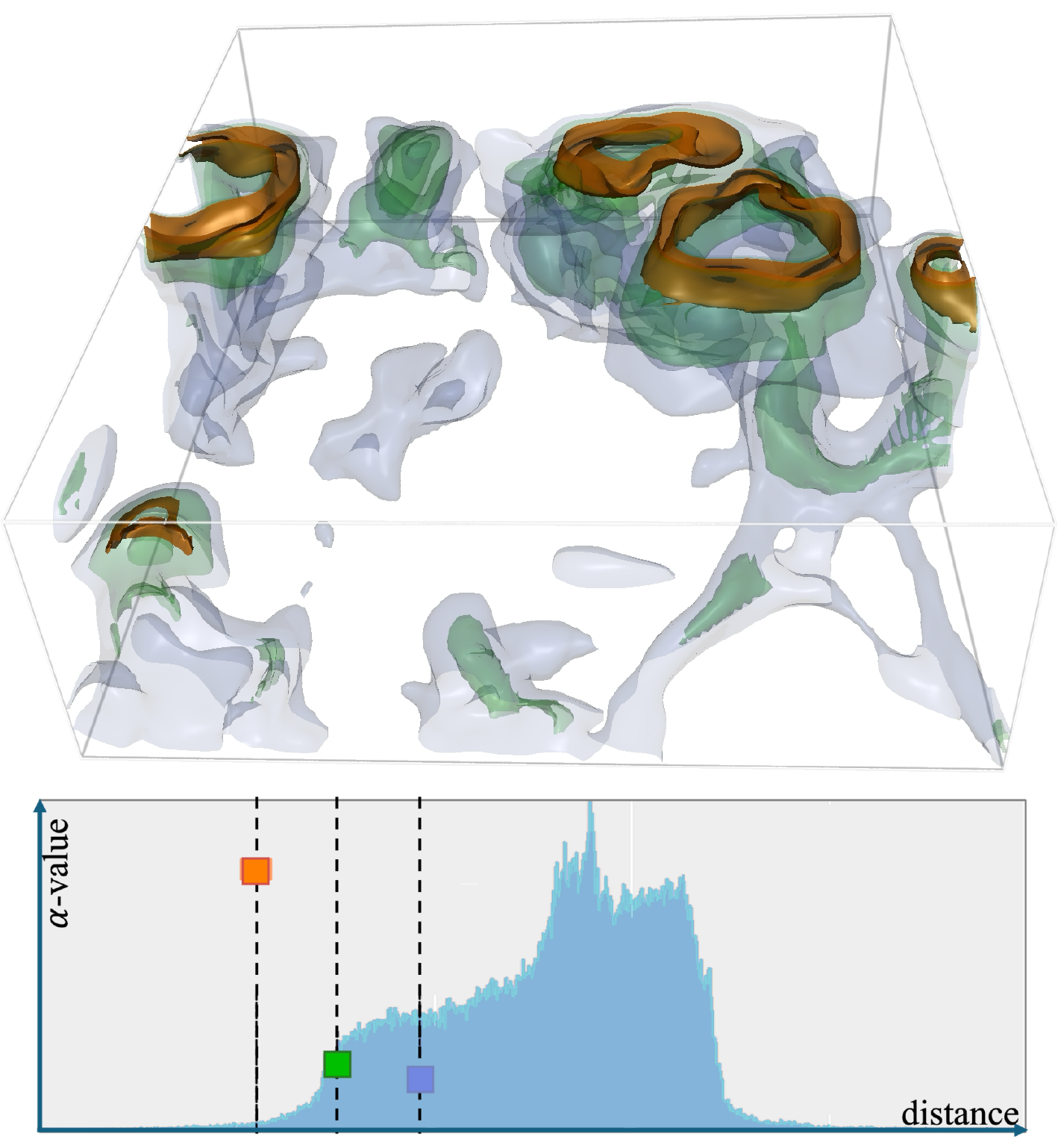}%
        \label{fig:subfig2}%
    }
    \hfill
    \subfloat[\small TIMT Segmentation]{%
        \includegraphics[width=0.3\textwidth]{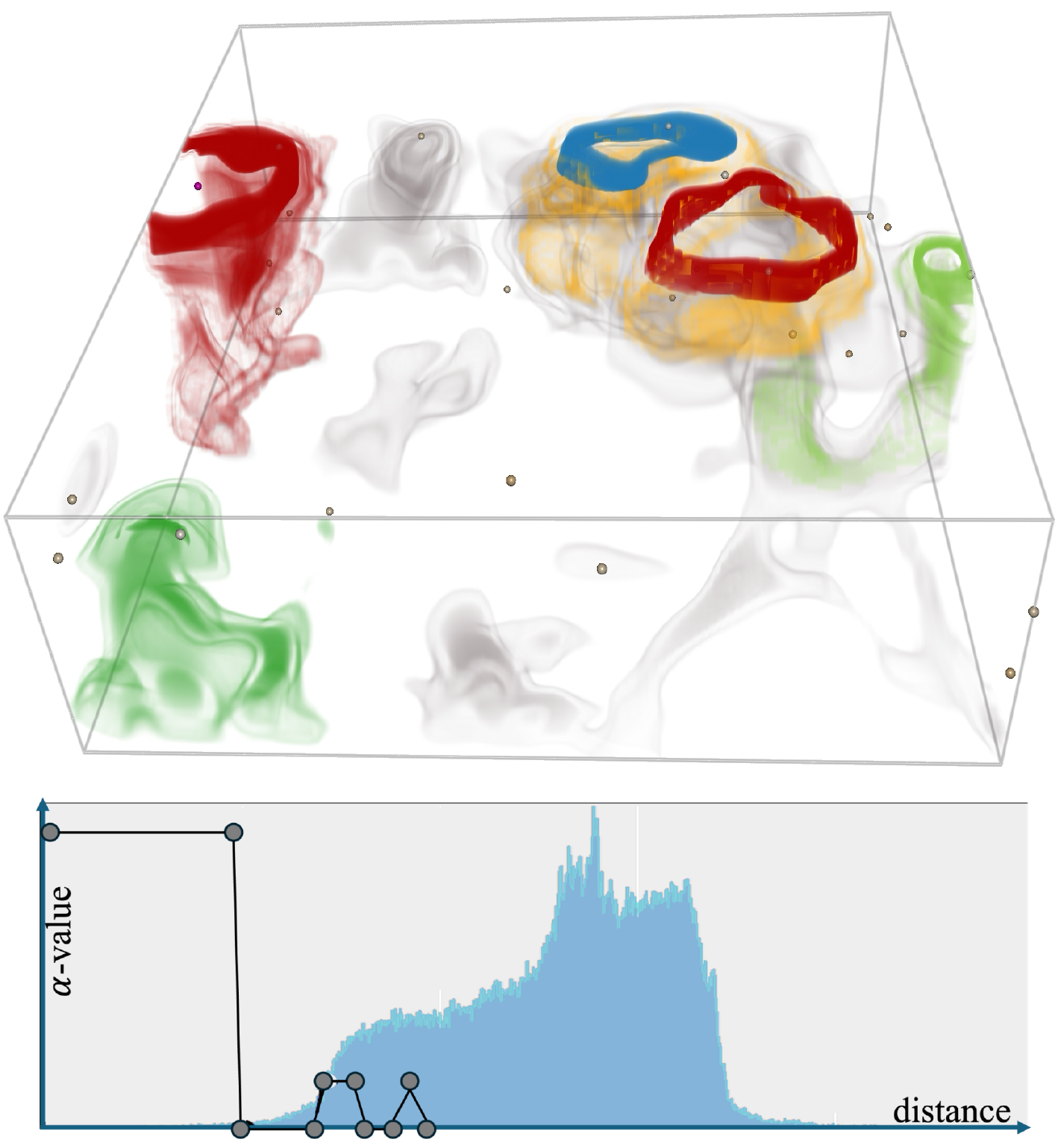}%
        \label{fig:subfig3}%
    }
    \caption{
        Images (a) and (b) demonstrate how the use of a polygon and a point trait \cmt{allow} for more flexibility also visualizing context to the areas of interest. As the distance to the trait is increased, we increase the allowed range of humidity and temperature, which sweeps downwards along the features, showing the different physical properties of different parts of the features. The use of a TIMT in (c) allows for the extraction of regions of interest with domain segmentation. The bottom row shows the corresponding distance distributions and transfer functions used.
    }
    \label{fig:cartesian-traits}
\end{figure*}

\begin{figure}
    \centering
       \includegraphics[width=0.4\linewidth]{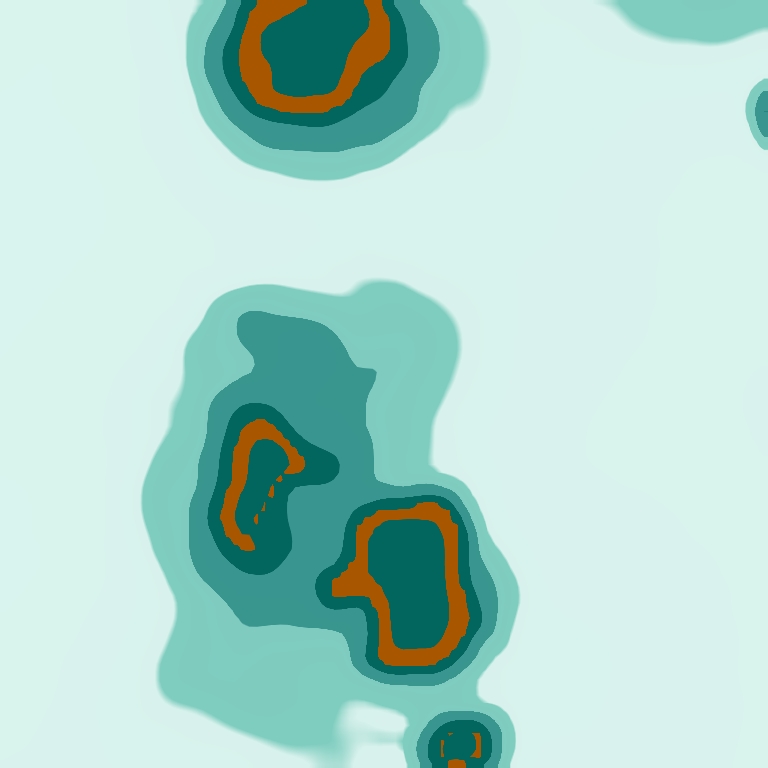} 
       \includegraphics[width=0.4\linewidth]{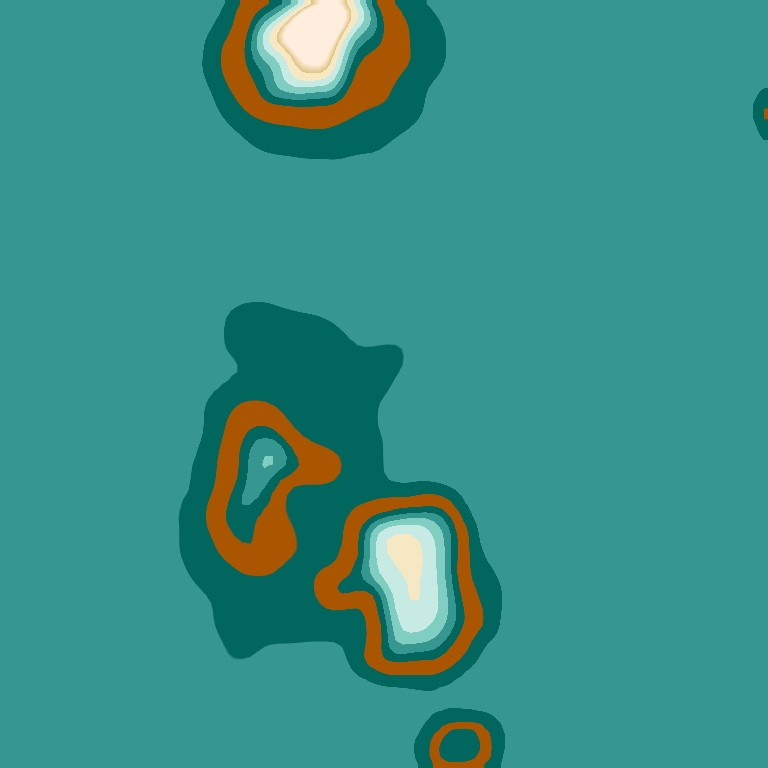}
    \caption{
       Closeup of a slice through the data set comparing the polygon trait (left) and the point trait (right). The structures of interest are highlighted in brown.}
    \label{fig:polygon-point}
\end{figure}

\subsection{Convective Cloud Formation}
The formation of convective clouds is difficult to model because resolving such a fine-scale process with numerical simulation is very computationally intensive. To improve climate and weather predictions, scientists are actively developing simplified models of how convective clouds are formed. One way to study convective cloud formation is to inject a tracer gas into numerical weather simulations~\cite{denby2022characterising, couvreux2010resolved}. 
This tracer gas tracks the bulk air movement from the Earth's surface through the boundary layer of the atmosphere where clouds are formed. 
By setting a threshold on the concentration of that tracer, scientists define cloud-triggering structures in the atmosphere as connected components of the tracer field's super-level sets. 
The challenge then is to analyze the physical properties of these structures, such as temperature and humidity.

We propose to study this multivariate data by using fiber surfaces, feature level sets, and TIMTs. 
Once scientists define the cloud-triggering structures by selecting a threshold for the tracer concentration, we can compute an isosurface and then simplify and color-code each \cmt{connected} component using the merge  tree~\cite{Carr2003a}. 
The volumes that represent the structures are then projected onto the continuous scatterplot of temperature and humidity, see~\fref{fig:cartesian-fs}(a). 
The scatterplot and projected structures guide the selection of fiber surface control polygons, allowing users to study certain properties of the structures and relate them to the spatial domain, see~\fref{fig:cartesian-fs}(c).
For example, examining areas with low humidity and high temperature.
However, as shown in~\fref{fig:cartesian-fs}(b), multiple isosurfaces and fiber surfaces can intersect and occlude each other. 
Cartesian fiber surfaces, proposed by Hristov~\cite{hristov_thesis_2022}, visualize only the parts of the cloud-triggering features with desired properties. 

Beyond Cartesian fiber surfaces, we can apply Cartesian traits and \cmt{TIMTs} to generate feature level sets. This approach offers greater flexibility, enabling us to examine the entire combined distance field with respect to a trait characterizing the properties of interest. Initially, we use the control polygon in a bivariate field of temperature and humidity, as used for the fiber surfaces in~\fref{fig:cartesian-fs}, combined with the trace concentration threshold as a Cartesian trait.
In~\fref{fig:cartesian-traits}(a), we present an isosurface volume rendering of the resulting feature level set for the zero level, shown in orange, which reproduces the fiber surface. Moving to higher levels provides the context for the wider distribution of temperature and humidity within the structures, depicted as blue and green transparent surfaces.

Similarly, in~\fref{fig:cartesian-traits}(b), we demonstrate the use of a point trait, positioned just above the polygon in the continuous scatterplot. As expected, near zero, the feature level set is empty. As we move away from zero, the features in the upper part, similar to the fiber surfaces, appear and develop like the polygon trait. 
\fref{fig:polygon-point} shows a slice comparing the polygon-based trait and the point trait feature level sets. Note that the range of distances in the field differs between the two settings, resulting in different color representations. The most interesting regions, corresponding to the original fiber surface and highlighted in brown, are very similar.

Finally, in~\fref{fig:cartesian-traits}\cmt{(c)}, we present the results of using the TIMT to obtain a segmentation of the domain into separate features. This method allows us to automatically detect changes in the combined distance field as we increase the distance from the point trait and intersect new structures. Notably, the TIMT captures all significant features automatically.

\section{Discussion and Conclusions}
\label{sec:conclusions}

In this paper, we present methods to enhance the application of Feature Level Sets (FLS). Our contributions include techniques to support trait design through the use of Cartesian traits, as well as the automatic suggestion of point traits using dictionary learning. Additionally, we introduce the Trait-Induced Merge Tree (TIMT), which facilitates feature selection by providing intuitive guidance. Finally, we enable users to interact with the merge tree by querying it with various methods and parameters, generating corresponding renderings to support analysis and visualization.

\paragraph*{Trait-Induced Merge Tree} With TIMTs we have introduced a topology-based segmentation of multi-field data based on traits.
They offer a novel approach to topological data analysis for multi-variate data and tensor fields by generalizing fundamental topological concepts, as demonstrated with the merge tree. Notions like persistence can be naturally extended to multi-fields or tensor fields, supporting multi-scale analysis. The concept of TIMTs is straightforward and easier to interpret compared to previously proposed methods for topological analysis of multi-fields. The computation of feature level sets is simple, and merge tree computation is well-researched and available through open-source libraries.

We demonstrated how TIMTs can be used to guide the exploration of multi-variate data by providing intuitive segmentation of the domain. They help to define, select, and interact with localized features whose properties are determined by the trait. The user interface supports browsing the segmentation via a legend or slice. However, determining parameters such as persistence thresholds or crown height can still be challenging without visual aids. 
A general observation when using feature level sets in visualization is that while the visualization effectively represents a specific trait, it does not convey detailed information about the individual fields. 

\paragraph*{Trait design} 
\cmt{In contrast to the computation of TIMTs, trait design is less straightforward; it can be tedious and involves several potential pitfalls. To address these challenges, we propose two complementary approaches to ease this process. Cartesian traits are designed for scenarios where domain experts already have a relatively clear understanding of their features and traits. They simplify the construction of complex traits by breaking them down into independent components, making the process more intuitive. Furthermore, it enhances the efficiency of distance field computations.
For cases where users are less familiar with the data and its structure, more automated, data-driven methods are essential. The automatic suggestion of point traits using dictionary learning is a promising step in this direction. While this approach shows potential, demonstrated in our case studies, the properties of the generated dictionaries remain not fully understood, necessitating critical evaluation of the results. There remain still lots of opportunities for improvement and further research.

A remaining key challenge in trait design is determining which attributes to include in the attribute $\Aspace$. Equally important is the task of specifying an appropriate metric for this space, particularly when the attributes have varying scales and units. These decisions are crucial yet non-trivial.}

\paragraph*{Computational aspects} 
Efficient algorithms for most steps in the pipeline are well-known and often available as open-source implementations. The FLS implementation used in this article has been integrated into the Inviwo framework~\cite{Jonsson2020b}. It is GPU-based and performed at interactive frame rates for all data sets used. The computation time of the merge tree depends on the complexity of the distance field; in our case studies, the longest computation took less than a minute (42.9 seconds). Feature extraction based on the methods presented in~~\sref{sec:query_methods} took a comparable amount of time. Additionally, the feature extraction process can be offloaded to the GPU and should be performed at interactive frame rates in workflows for domain experts.
However, there is one critical aspect related to interpolation: some derived attributes, such as eigenvalues for tensor fields and the pull-back of the distance field, are non-linear in the domain. This can lead to artifacts when interpolating the per-computed values at the vertices. To compute exact level sets, the distance field would have to be evaluated during the ray-marching process for every evaluation. Similar problems might arise in the computation of the merge tree, which typically assumes a piecewise linear behavior of the data in the domain~\cite{BinMasood2021b}.

\paragraph*{Case studies} The proposed method has been demonstrated in several case studies with diverse characteristics, consistently producing the expected and desired outcomes.
In Case Study B, within the domain of tensor fields, the method enabled the computation of features similar to 'tensor isosurfaces,' fulfilling a long-standing request from our collaboration partners.
For Case Study C, applying TIMTs to data representing molecular electronic transitions results in the automatic identification of donor and acceptor regions in a molecule using a simple point trait. This result was highly appreciated by domain scientists, as no other analysis methods are available for this task. This case study also demonstrates the strength of point traits, which greatly simplify the analysis compared to previous work using fiber surfaces, where specific control polygons must be set up for each dataset~\cite{sharma2023continuous}.
Feature level sets are generally less sensitive to small changes in the feature. This can be observed in the convective cloud formation case study E, where the use of a point feature produces very similar results to a control polygon designed to use fiber surfaces.
Case Study D, which deals with vortex re-connection, exemplifies the manual configuration of a trait to extract a complex structure. Moreover, the study demonstrates that dictionary learning also has the potential to identify and extract features of interest. The method suggested a trait similar to the manually defined one. However, it's important to note that atoms do not necessarily have semantic meaning and it can be difficult to interpret the connection between atoms and features. Using parallel coordinates to visualize the properties of atoms can aid in this regard. In Case Study A, we attempt to address this limitation by using TIMT to explore the atoms of learned dictionaries. By selecting a simple dataset with well-known features, we could compare dictionaries learned under different parameters. This approach reveals significant potential for enhancing and guiding the dictionary learning process in various contexts, such as introducing constraints on atoms. A detailed analysis of atoms could provide insights into finding a balance between representation accuracy and interpretability, as well as determining the optimal number of atoms needed to capture the underlying structure without overfitting.

\section*{acknowledgments}
{This work was supported by the Wallenberg AI, Autonomous Systems and Software Program (WASP) funded by the Knut and Alice Wallenberg Foundation, the SeRC (Swedish e-Science Research Center), the ELLIIT environment for strategic research in Sweden, and the Swedish Research Council~(VR) grants 2019-05487 and 2023-04806. The authors thank Mathieu Linares (KTH Royal Institute of Technology) for the charge transfer simulation data and valuable feedback on the results of the case study.  The flow dataset was provided by Professor Jan Nordstr\"om (Link\"oping University) and Dr. Marco Kupiainen (Rossby Centre, SMHI).}
\begin{appendices}
\section{Note on the stability of TIMTs}
\label{appendix_A}
Morozov et al.~\cite{ Morozov2013b} have proven the stability of merge trees, with respect to an interleaving distance, to perturbations of the functions that define them. 
In Theorem 2 (Stability) they state that for two scalar functions 
$f,g:\Xspace\rightarrow\Rspace$ 
and their corresponding merge trees $\text{MT}_f$ and $\text{MT}_g$, the interleaving distance between the trees does not exceed the largest difference between the two functions 
$d_I(\text{MT}_f,\text{MT}_g) \le \sup_{x\in \Xspace}|f(x)-g(x)|$. 
In our case, the two merge trees are the TIMTs of two induced distance fields $h_T$ and $h_{T^\prime}$ for two different traits $T$ and its respective $T^\prime$. To prove the stability of the TIMTs we have to prove the stability of the field-induced distance fields with respect to changes in the trait. 
For point traits, we can show that the changes in the distance field are limited by the distance between the two traits, as a consequence of the triangle inequality in $\Aspace$.
Let $T$ and $T^\prime$ be two point traits in $\Aspace$, $h_T$ and $h_{T^\prime}$ the corresponding trait induced distance fields $\Xspace\rightarrow\Rspace$ and $x$ an arbitrary poiont in $\Xspace$:
\begin{equation}
|h_T(x)-h_{T^\prime}(x)| = |d_{\Aspace}(f(x), T)- d_{\Aspace}(f(x), T^\prime)|\le d_{\Aspace}(T,T^\prime).
\label{eq:triangle}
\end{equation}
We can directly follow from~\cite{ Morozov2013b} and~\eqref{eq:triangle}:

\begin{equation}
d_I(\text{TIMT}_T,\text{TIMT}_{T^\prime}) \le 
\sup_{x\in\Xspace} |h_T(x)-h_{T^\prime}(x)|\le d_{\Aspace}(T,T^\prime).
\end{equation}

This shows that the changes in the TIMTs regarding the interleaving distance are constrained by the distance of the point trait.
More generally, for any two traits $T, T^\prime \subseteq \Aspace$, the interleaving distance between the two merge trees is limited by differences in the two distance fields $d_T$ and $d_{T{^\prime}}$:
\begin{equation}
\begin{split}
d_I(\text{TIMT}_T,\text{TIMT}_{T^\prime}) \le 
\sup_{x\in\Xspace} |h_T(x)-h_{T^\prime}(x)|\\
= \sup_{x\in\Xspace} |d_T(f(x)) - d_{T^\prime}(f(x))|
= \sup_{a \in f(\Xspace)} |d_T(a) - d_{T^\prime}(a)|\\
\le \sup_{a \in \Aspace} |d_T(a) - d_{T^\prime}(a)|.
\end{split}
\end{equation}
The first line follows from~\cite{Morozov2013b}; the second expands the definition of trait-induced distance fields and then maps the expression from $\Xspace$ to $\Aspace$ via $f$; the final line uses the standard inequality from real analysis $\sup(A) \leq \sup(B)$ for $A \subseteq B$.

\cmt{Since Hausdorff distance $d_H$ between the two traits $T$ and $T'$ in the metric space $\Aspace$ is equivalent to $\sup_{a \in \Aspace} |d_T(a) - d_{T^\prime}(a)|$, we finally have the following stability result:

\begin{equation}
d_I(\text{TIMT}_T,\text{TIMT}_{T^\prime}) \le d_H(T, T').
\end{equation}
}

\newpage
\section{Experiments with Phantom HARDI Data for Dictionary Learning}
\label{appendix_B}
Determining the optimal number of atoms for a dictionary is a challenging task, as it requires balancing interpretability, meaningful and effective representation, and the risk of redundancy and overfitting. This is similar to setting an appropriate number of clusters for data representation. The number of atoms in a dictionary directly affects its ability to capture the underlying structure of the data. Too few atoms may result in a simplified structure, failing to capture important details, while too many atoms can lead to redundancy. 

The phantom data used in the case study,  
described in Sec. VII of the main paper, is high-dimensional but contains a simple underlying structure, which makes it suitable to explore how different dictionary sizes affect its representation quality. To find the most effective dictionary for representing this phantom data, we experiment with various dictionary sizes. (Dictionary $D_{B}$ consists of 30 atoms, dictionary $D_{C}$ of 10 atoms, and dictionary $D_{A}$ of 6 atoms.)
We evaluated the performance of each dictionary size based on how well it captures the essential directional features in the data. 

\begin{figure*}[htb]
    \centering  
    \subfloat[\label{subfig:dis_dictionary_A} \small heatmap for $D_A$]{
        \includegraphics[width=0.31\textwidth]{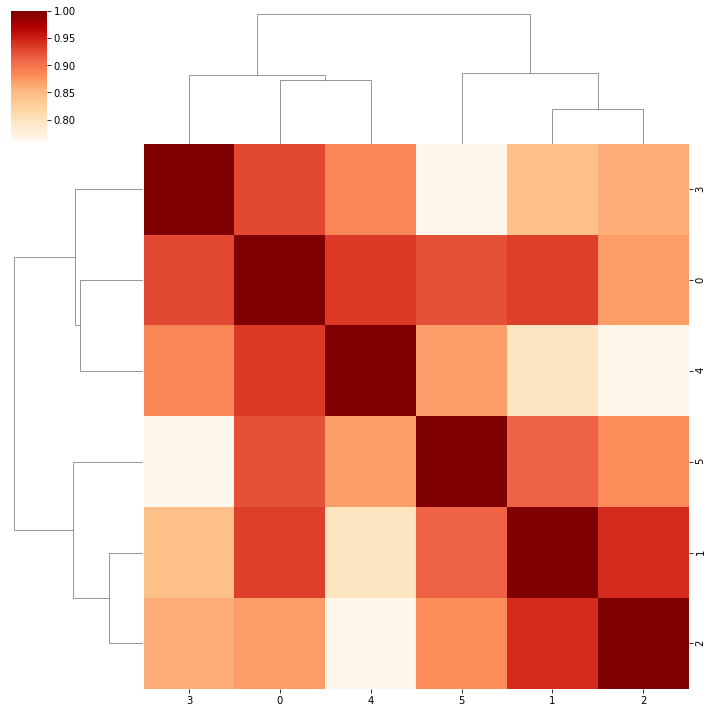}}
    \hfill
    \subfloat[\label{subfig:dis_dictionary_B}\small heatmap for $D_B$]{
        \includegraphics[width=0.31\textwidth]{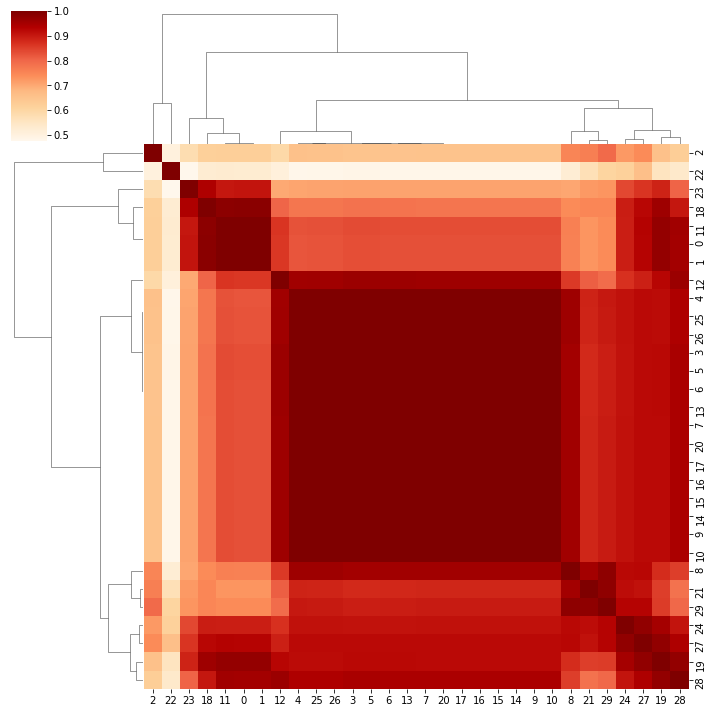}}
    \hfill
    \subfloat[\label{subfig:dis_dictionary_C}\small heatmap for $D_C$]{
        \includegraphics[width=0.31\textwidth]{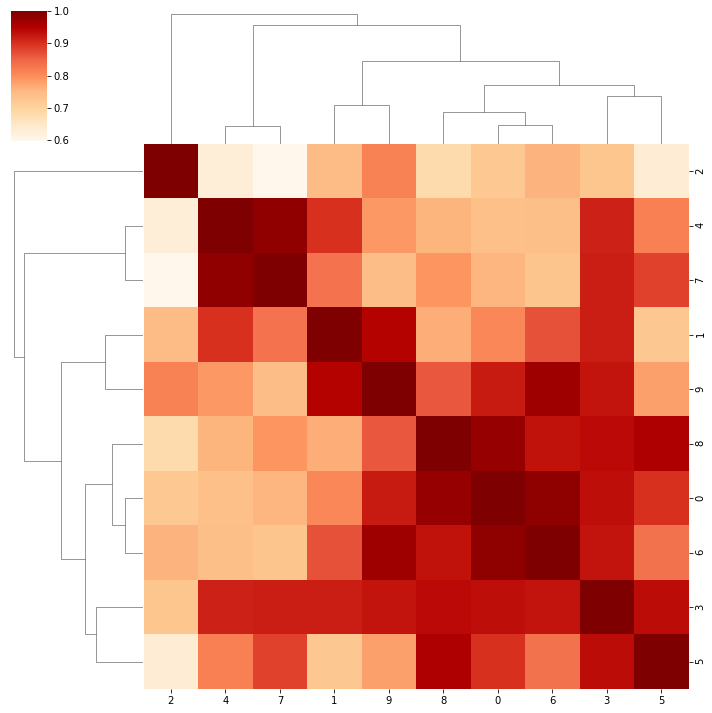}}
    \hfill
    \caption{Clustered heatmaps of distance matrix from 3 different dictionaries: (a) $D_{A}$ with 6 atoms, (b) $D_{B}$ with 30 atoms, and (c) $D_{C}$ with 10 atoms.}
    \label{fig:clustered_map}   
\end{figure*}

To guide the selection of atoms for our experiments and gain a better understanding of their relationships, we begin by computing pairwise similarities between atoms for each dictionary. This helps group atoms with similar properties into clusters. The results are presented as clustered heatmaps in Fig.~\ref{fig:clustered_map}.
In the heatmaps, we can clearly observe that dictionaries $D_{B}$ and $D_{c}$ exhibit distinct clusters with high similarity, which is not the case for dictionary $D_{A}$.
To verify this observation, we first selected two atoms from the largest cluster in dictionary $D_{B}$ (depicted as the large red square in Fig.~\ref{fig:clustered_map}(b). 

\begin{figure*}[htb]
    \includegraphics[width=0.95\textwidth]{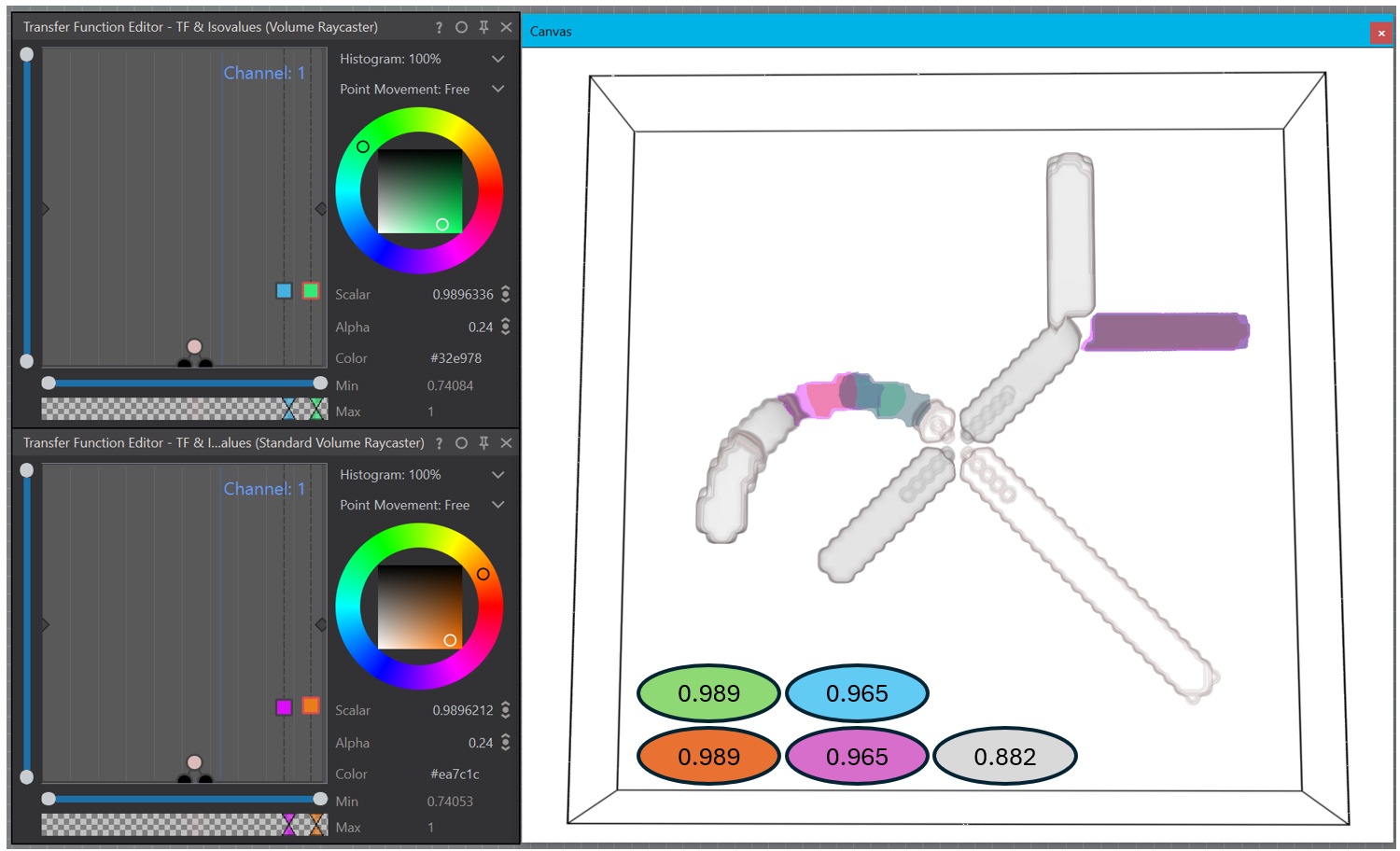}
    \caption{An example of overlapping representation for atoms 9 and 10 in $D_{B}$. Transfer functions (left): high similarity values for atom 9 are highlighted in green and blue, while high similarity values for atom 10 are highlighted in orange and magenta. Volume rendering (right): regions with the highest similarity to the two atoms have significant overlap suggesting redundancy in the dictionary $D_{B}$. }
   \label{fig:vr_atom_dif}
    
\end{figure*}

These atoms within the same cluster tend to represent similar features, leading to overlapping FLS representations, as illustrated in Fig.~\ref{fig:vr_atom_dif}. The renderings in Fig.~\ref{fig:vr_atom_dif} are generated using the same transfer function settings, but they are assigned different color schemes to emphasize the overlap and differences between the highlighted regions associated with these two atoms. Although most parts of the FLS overlap, there are some differences in the curved part of the structure.

\begin{figure*}[htb]
    \centering  
    \subfloat[\label{subfig:vr_atom2} \small Atom-trait 2, $D_B$]{
        \includegraphics[width=0.315\textwidth]{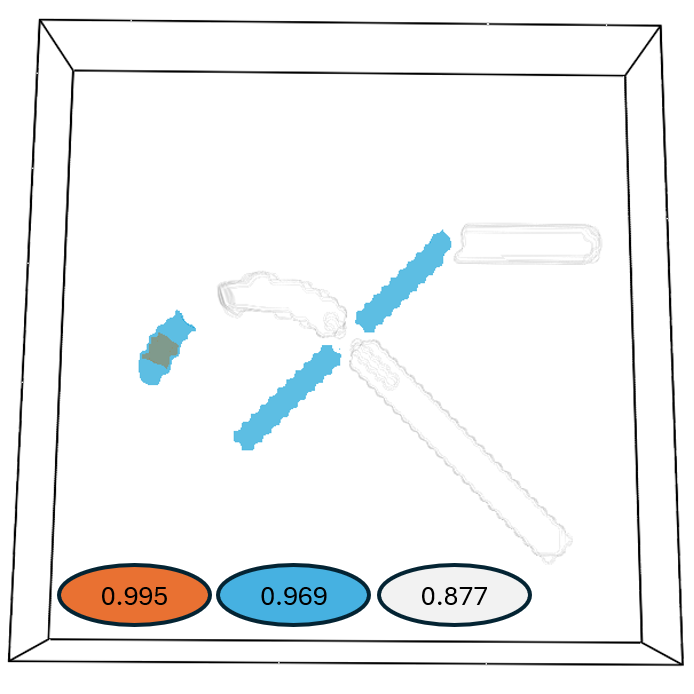}}
    \hfill
    \subfloat[\label{subfig:vr_atom22}\small Atom-trait 22, $D_B$]{
        \includegraphics[width=0.316\textwidth]{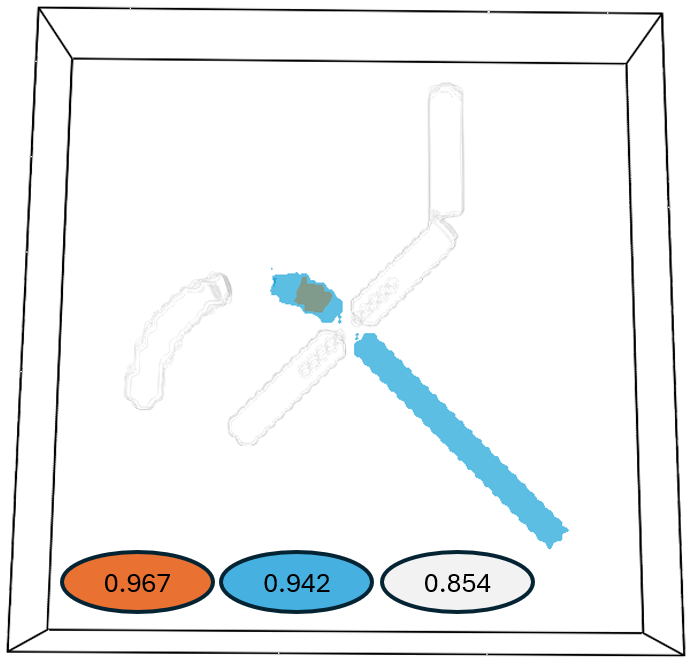}}
    \hfill
    \subfloat[\label{subfig:vr_atom11}\small Atom-trait 11, $D_B$]{
        \includegraphics[width=0.312\textwidth]{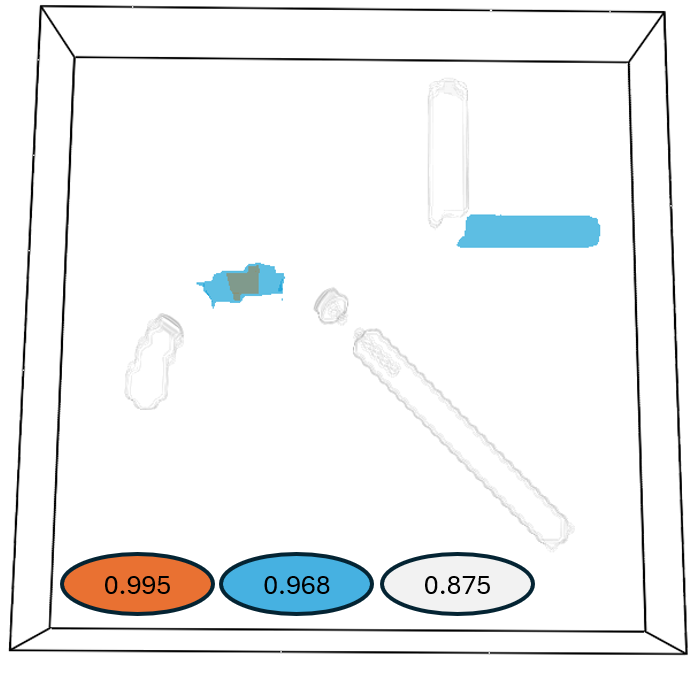}}
    \hfill  
    \\
    \subfloat[\label{subfig:vr_atom24}\small Atom-trait 24, $D_B$]{
        \includegraphics[width=0.32\textwidth]{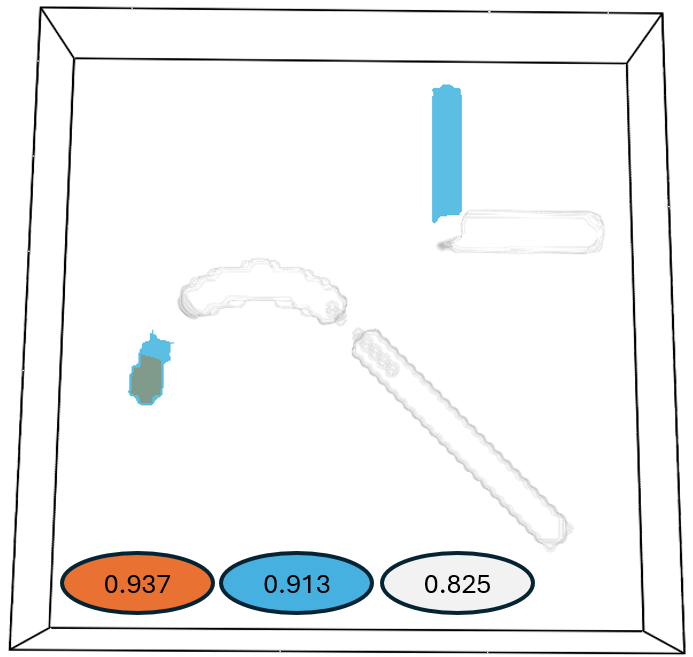}}
    \hfill
    \subfloat[\label{subfig:vr_atom28}\small Atom-trait 28, $D_B$]{
        \includegraphics[width=0.315\textwidth]{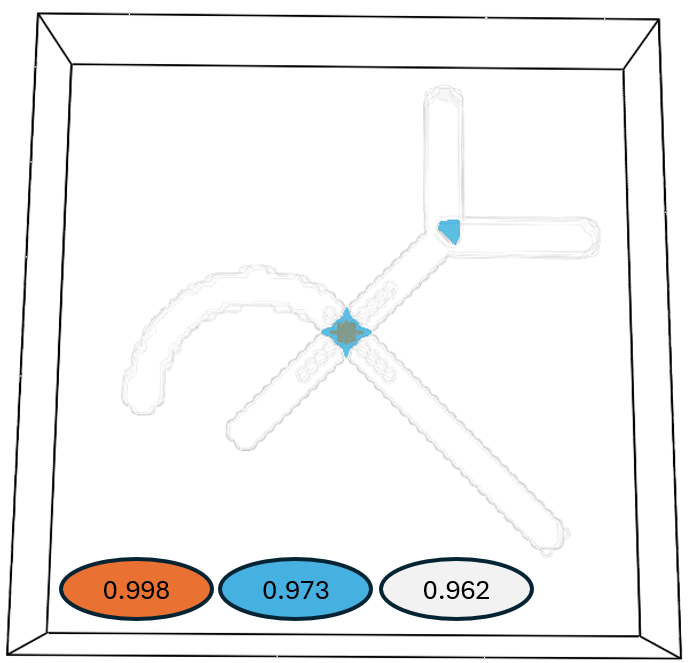}}
    \hfill
    \subfloat[\label{subfig:vr_atom29} \small Atom-trait 29, $D_B$]{
       \includegraphics[width=0.313\textwidth]{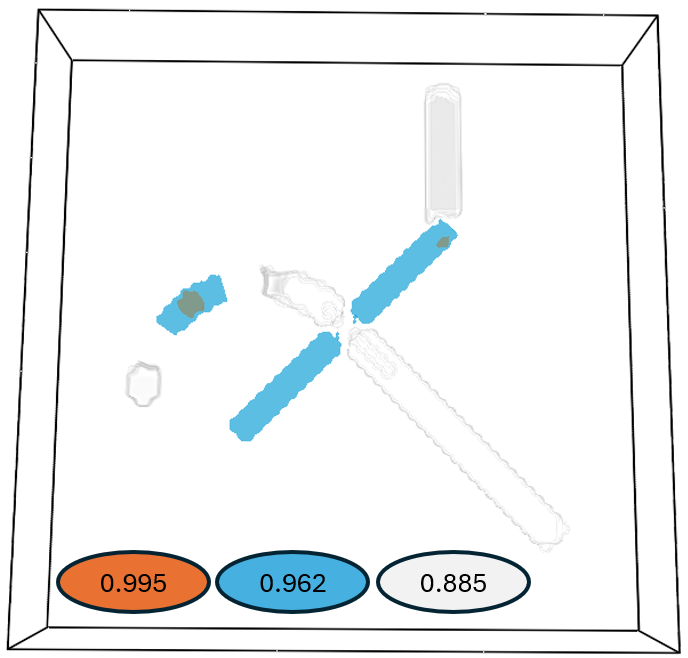}}
    \hfill
\caption{Volume renderings of highlighted regions with the highest similarity corresponding to representative atoms in $D_{B}$ from different clusters. Iso-values for each transfer function are shown in colored ovals. Highlighted regions from (a) to (e) collectively represent the full structure of the phantom data. Specifically, regions from (a) through (d) represent different sections of the phantom data, each oriented along a specific direction, while region (e) captures the intersecting areas. Additionally, (a) and (f) show that atoms 2 and 29 represent similar areas but belong to different clusters.}   
\label{fig:vr}
\end{figure*}

In the next step, we take one representative atom from all remaining clusters in $D_{B}$,  presented in Fig.~\ref{fig:vr}. The chosen transfer functions in all these images are very similar. Looking at the set of six images we can see that the entire structure is well captured.
We obtained similar results for the dictionary $D_{C}$ with 10 atoms. The results for the smallest dictionary are shown in the results of the main paper. While these also capture most structures well, they miss the probably most interesting part that represents the fiber crossings represented by atom 28 in $D_{B}$. 

These examples show how FLS can contribute to the exploration and understanding of sparse dictionaries.

\clearpage

\clearpage

\end{appendices}
\clearpage
\newpage

\bibliographystyle{IEEEtran}
\bibliography{trait.bib}

\begin{thebibliography}{10}
\providecommand{\url}[1]{#1}
\csname url@samestyle\endcsname
\providecommand{\newblock}{\relax}
\providecommand{\bibinfo}[2]{#2}
\providecommand{\BIBentrySTDinterwordspacing}{\spaceskip=0pt\relax}
\providecommand{\BIBentryALTinterwordstretchfactor}{4}
\providecommand{\BIBentryALTinterwordspacing}{\spaceskip=\fontdimen2\font plus
\BIBentryALTinterwordstretchfactor\fontdimen3\font minus \fontdimen4\font\relax}
\providecommand{\BIBforeignlanguage}[2]{{%
\expandafter\ifx\csname l@#1\endcsname\relax
\typeout{** WARNING: IEEEtran.bst: No hyphenation pattern has been}%
\typeout{** loaded for the language `#1'. Using the pattern for}%
\typeout{** the default language instead.}%
\else
\language=\csname l@#1\endcsname
\fi
#2}}
\providecommand{\BIBdecl}{\relax}
\BIBdecl

\bibitem{sharma2023continuous}
M.~Sharma, T.~B. Masood, S.~S. Thygesen, M.~Linares, I.~Hotz, and V.~Natarajan, ``Continuous scatterplot operators for bivariate analysis and study of electronic transitions,'' \emph{IEEE Transactions on Visualization and Computer Graphics}, vol.~30, no.~7, pp. 3532--3544, 2024.

\bibitem{Jankowai2020a}
J.~Jankowai, R.~Sk{\aa}nberg, D.~J{\"o}nsson, A.~Ynnerman, and I.~Hotz, ``{T}ensor {V}olume {E}xploration {U}sing {A}ttribute {S}pace {R}epresentatives,'' in \emph{LEVIA 2020: Leipzig Symposium on Visualization in Applications}, 2020.

\bibitem{JankowaiHotz2019}
J.~{Jankowai} and I.~{Hotz}, ``{F}eature {L}evel-{S}ets: {G}eneralizing iso-surfaces to multi-variate data,'' \emph{IEEE Transactions on Visualization and Computer Graphics}, vol.~26, no.~2, pp. 1308--1319, Feb 2020.

\bibitem{fibersurfaces}
H.~Carr, Z.~Geng, J.~Tierny, A.~Chattopadhyay, and A.~Knoll, ``{F}iber {S}urfaces: {G}eneralizing isosurfaces to bivariate data,'' \emph{{Computer Graphics Forum}}, vol.~34, no.~3, pp. 241--250, Jun. 2015.

\bibitem{Sane2021}
S.~Sane, T.~M. Athawale, and C.~R. Johnson, ``Visualization of uncertain multivariate data via feature confidence level-sets,'' in \emph{EuroVis - Short Papers}, M.~Agus, C.~Garth, and A.~Kerren, Eds.\hskip 1em plus 0.5em minus 0.4em\relax Eurographics Association, 2021.

\bibitem{Nguyen2021a}
D.~B. Nguyen, R.~O. Monico, and G.~Chen, ``{A} {V}isualization {F}amework for {M}ulti-{S}cale {C}oherent {S}tructures in {T}aylor-{C}ouette {T}urbulence,'' \emph{{IEEE Transactions on Visualization and Computer Graphics}}, vol.~27, no.~2, pp. 902--912, 2021.

\bibitem{Lei2024}
D.~Lei, E.~Miandji, J.~Unger, and I.~Hotz, ``{Sparse q-ball imaging towards efficient visual exploration of HARDI data},'' \emph{{Computer Graphics Forum}}, vol.~43, no.~3, p. e15082, 2024.

\bibitem{Weber2007a}
G.~Weber, S.~Dillard, H.~Carr, V.~Pascucci, and B.~Hamann, ``{T}opology-{C}ontrolled {V}olume {R}endering,'' \emph{IEEE Transactions on Visualization and Computer Graphics}, vol.~13, pp. 330--41, 04 2007.

\bibitem{Nilsson2022}
E.~Nilsson, J.~Lukasczyk, W.~Engelke, T.~B. Masood, G.~Svensson, R.~Caballero, C.~Garth, and I.~Hotz, ``Exploring cyclone evolution with hierarchical features,'' in \emph{2022 Topological Data Analysis and Visualization (TopoInVis)}.\hskip 1em plus 0.5em minus 0.4em\relax IEEE, 2022, pp. 92--102.

\bibitem{jankowai2023multi}
J.~Jankowai, T.~B. Masood, and I.~Hotz, ``Multi-field visualisation via trait-induced merge trees,'' in \emph{2023 Topological Data Analysis and Visualization (TopoInVis)}.\hskip 1em plus 0.5em minus 0.4em\relax IEEE, 2023, pp. 21--29.

\bibitem{Roberts2007}
J.~C. Roberts, ``{S}tate of the {A}rt: {C}oordinated multiple views in exploratory visualization,'' in \emph{International Conf. on Coordinated and Multiple Views in Exploratory Visualization}, 2007, pp. 61--71.

\bibitem{Ljung2016}
P.~Ljung, J.~Kr{\"u}ger, E.~Gr\"oller, M.~Hadwiger, C.~D. Hansen, and A.~Ynnerman, ``{S}tate of the {A}rt in {T}ransfer {F}unctions for {D}irect {V}olume {R}endering,'' in \emph{Computer Graphics Forum}, vol.~35, no.~3, 2016, pp. 669--691.

\bibitem{Wang2012}
Y.~Wang, J.~Zhang, D.~J. Lehmann, H.~Theisel, and X.~Chi, ``{A}utomating {T}ransfer {F}unction {D}esign {W}ith {V}alley {C}ell-{B}ased {C}lustering of 2{D} {D}ensity {P}lots,'' \emph{Computer Graphics Forum}, vol.~31, no. 3pt4, pp. 1295--1304, 2012.

\bibitem{Cai2017}
L.~Cai, B.~P. Nguyen, C.-K. Chui, and S.-H. Ong, ``{A} {T}wo-{L}evel {C}lustering {A}pproach for {M}ultidimensional {T}ransfer {F}unction {S}pecification in {V}olume {V}isualization,'' \emph{The Visual Computer}, vol.~33, pp. 163--177, 2017.

\bibitem{Dobrev2011}
P.~Dobrev, T.~van Long, and L.~Linsen, ``{A} {C}luster {H}ierarchy-{B}ased {V}olume {R}endering {A}pproach for {I}nteractive {V}isual {E}xploration of {M}ulti-{V}ariate {V}olume {D}ata,'' in \emph{Proceedings of Vision, Modeling, and Visualization (VMV'11)}, 2011, pp. 137--144.

\bibitem{vanKreveld1997a}
M.~van Kreveld, R.~van Oostrum, C.~Bajaj, V.~Pascucci, and D.~Schikore, ``{C}ontour {T}rees and {S}mall {S}eed {S}ets for {I}sosurface {T}raversal,'' in \emph{Proceedings of the Thirteenth Annual Symposium on Computational Geometry}, ser. SCG '97.\hskip 1em plus 0.5em minus 0.4em\relax New York, NY, USA: Association for Computing Machinery, 1997, pp. 212--220.

\bibitem{Bremer2011}
P.-T. Bremer, G.~Weber, J.~Tierny, V.~Pascucci, M.~Day, and J.~Bell, ``{I}nteractive {E}xploration and {A}nalysis of {L}arge-{S}cale {S}imulations {U}sing {T}opology-{B}ased {D}ata {S}egmentation,'' \emph{IEEE Transactions on Visualization and Computer Graphics}, vol.~17, no.~9, pp. 1307--1324, 2011.

\bibitem{Bock2018}
A.~Bock, H.~Doraiswamy, A.~Summers, and C.~Silva, ``{T}opoangler: {I}nteractive topology-based extraction of fishes,'' \emph{IEEE Transactions on Visualization and Computer Grapics}, vol.~24, no.~1, pp. 812--821, 2018.

\bibitem{Felix}
N.~Shivashankar, P.~Pranav, V.~Natarajan, R.~v.~d. Weygaert, E.~P. Bos, and S.~Rieder, ``{F}elix: {A} topology based framework for visual exploration of cosmic filaments,'' \emph{IEEE Transactions on Visualization and Computer Graphics}, vol.~22, no.~6, pp. 1745--1759, Jun 2016.

\bibitem{topologystar}
C.~Heine, H.~Leitte, M.~Hlawitschka, F.~Iuricich, L.~De~Floriani, G.~Scheuermann, H.~Hagen, and C.~Garth, ``{A} {S}urvey of {T}opology-{B}ased {M}ethods in {V}isualization,'' \emph{Computer Graphics Forum}, vol.~35, no.~3, pp. 643--667, 2016.

\bibitem{Wu2017a}
K.~Wu, A.~Knoll, B.~J. Isaac, H.~Carr, and V.~Pascucci, ``{D}irect {M}ultifield {V}olume {R}ay {C}asting of {F}iber {S}urfaces,'' \emph{IEEE Transactions on Visualization and Computer Graphics}, vol.~23, no.~1, pp. 941--949, 2017.

\bibitem{Klacansky2017a}
P.~Klacansky, J.~Tierny, H.~Carr, and Z.~Geng, ``{F}ast and {E}xact {F}iber {S}urfaces for {T}etrahedral {M}eshes,'' \emph{IEEE Transactions on Visualization and Computer Graphics}, vol.~23, no.~7, pp. 1782--1795, 2017.

\bibitem{raithsalamislice1}
F.~Raith, C.~Blecha, T.~Nagel, F.~Parisio, O.~Kolditz, F.~G\"{u}nther, M.~Stommel, and G.~Scheuermann, ``{T}ensor {F}ield {V}isualization {U}sing {F}iber {S}urfaces of {I}nvariant {S}pace,'' \emph{IEEE Transactions on Visualization and Computer Graphics}, vol.~25, no.~1, pp. 1122--1131, Jan 2019.

\bibitem{raithsalamislice2}
C.~Blecha, F.~Raith, A.~J. Pr{\"a}ger, T.~Nagel, O.~Kolditz, J.~Ma{\ss}mann, N.~R{\"o}ber, M.~B{\"o}ttinger, and G.~Scheuermann, ``{F}iber {S}urfaces for {M}any {V}ariables,'' \emph{Computer Graphics Forum}, vol.~39, no.~3, pp. 317--329, 2020.

\bibitem{Athawale2021a}
T.~M. Athawale, B.~J. Stanislawski, S.~Sane, and C.~R. Johnson, ``{V}isualizing {I}nteractions {B}etween {S}olar {P}hotovoltaic {F}arms and the {A}tmospheric {B}oundary {L}ayer,'' in \emph{Proceedings of the Twelfth ACM International Conference on Future Energy Systems}, ser. e-Energy '21.\hskip 1em plus 0.5em minus 0.4em\relax New York, NY, USA: Association for Computing Machinery, 2021, pp. 377--381.

\bibitem{Carr2003a}
H.~Carr, J.~Snoeyink, and U.~Axen, ``{C}omputing {C}ontour {T}rees in all {D}imensions,'' \emph{Computational Geometry}, vol.~24, no.~2, pp. 75--94, 2003.

\bibitem{Cohen-SteinerEdelsbrunnerHarer2010}
D.~Cohen-Steiner, H.~Edelsbrunner, J.~Harer, and Y.~Mileyko, ``{L}ipschitz {F}unctions {H}ave $l_p$-{S}table {P}ersistence,'' \emph{Foundations of Computational Mathematics}, vol.~10, no.~2, pp. 127--139, 2010.

\bibitem{Pascucci2005a}
V.~Pascucci, K.~Cole-McLaughlin, and G.~Scorzelli, ``{M}ulti-{R}esolution {C}omputation and {P}resentation of {C}ontour {T}rees,'' in \emph{IASTED conference on Visualization, Imaging, and Image Processing (VIIP 2004)}, 2005, pp. 452--290.

\bibitem{hristov_thesis_2022}
\BIBentryALTinterwordspacing
P.~G. Hristov, ``\BIBforeignlanguage{en}{Hypersweeps, {Convective} {Clouds} and {Reeb} {Spaces}},'' Ph.D. dissertation, University of Leeds, Jun. 2022. [Online]. Available: \url{https://etheses.whiterose.ac.uk/31965/}
\BIBentrySTDinterwordspacing

\bibitem{frisken_2006}
S.~F. Frisken and R.~N. Perry, ``Designing with distance fields,'' in \emph{ACM SIGGRAPH 2006 Courses}, ser. SIGGRAPH '06.\hskip 1em plus 0.5em minus 0.4em\relax New York, NY, USA: Association for Computing Machinery, 2006, p. 60–66.

\bibitem{elad2010sparse}
M.~Elad, \emph{Sparse and redundant representations: from theory to applications in signal and image processing}.\hskip 1em plus 0.5em minus 0.4em\relax Springer Science \& Business Media, 2010.

\bibitem{Jonsson2020b}
D.~J{\"o}nsson, P.~Steneteg, E.~Sund\'en, R.~Englund, S.~Kottravel, M.~Falk, A.~Ynnerman, I.~Hotz, and T.~Ropinski, ``{Inviwo - A Visualization System with Usage Abstraction Levels},'' \emph{{IEEE Transactions on Visualization and Computer Graphics}}, vol.~26, no.~11, pp. 32--3254, 2020.

\bibitem{descoteaux2007regularized}
M.~Descoteaux, E.~Angelino, S.~Fitzgibbons, and R.~Deriche, ``Regularized, fast, and robust analytical q-ball imaging,'' \emph{Magnetic Resonance in Medicine: An Official Journal of the International Society for Magnetic Resonance in Medicine}, vol.~58, no.~3, pp. 497--510, 2007.

\bibitem{phantom}
A.~Leemans, B.~Jeurissen, J.~Sijbers, and D.~Jones, ``Explore{DTI}: a graphical toolbox for processing, analyzing, and visualizing diffusion mr data,'' in \emph{Proc. International Society for Magnetic Resonance in Medicine}, vol.~17, 2009, p. 3537.

\bibitem{leemans2005mathematical}
A.~Leemans, J.~Sijbers, M.~Verhoye, A.~Van~der Linden, and D.~Van~Dyck, ``Mathematical framework for simulating diffusion tensor mr neural fiber bundles,'' \emph{Magnetic Resonance in Medicine: An Official Journal of the International Society for Magnetic Resonance in Medicine}, vol.~53, no.~4, pp. 944--953, 2005.

\bibitem{Westin1997}
C.-F. Westin, S.~Peled, H.~Gudbjartsson, R.~Kikinis, and F.~A. Jolesz, ``Geometrical diffusion measures for {MRI} from tensor basis analysis,'' in \emph{Proc. International Society for Magnetic Resonance in Medicine '97}, Vancouver Canada, April 1997, p. 1742.

\bibitem{Masood2021Transitions}
T.~B. Masood, S.~S. Thygesen, M.~Linares, A.~I. Abrikosov, V.~Natarajan, and I.~Hotz, ``Visual analysis of electronic densities and transitions in molecules,'' \emph{Computer Graphics Forum}, vol.~40, no.~3, pp. 287--298, 2021.

\bibitem{sharma2021segmentation}
M.~Sharma, T.~B. Masood, S.~S. Thygesen, M.~Linares, I.~Hotz, and V.~Natarajan, ``Segmentation driven peeling for visual analysis of electronic transitions,'' in \emph{2021 {IEEE} Visualization Conference, {IEEE} {VIS} 2021 - Short Papers}.\hskip 1em plus 0.5em minus 0.4em\relax {IEEE}, 2021, pp. 96--100.

\bibitem{Bachthaler2008CSP}
S.~Bachthaler and D.~Weiskopf, ``Continuous scatterplots,'' \emph{IEEE Transactions on Visualization and Computer Graphics}, vol.~14, no.~6, pp. 1428--1435, 2008.

\bibitem{Gunther2018Vortex}
T.~G{\"{u}}nther and H.~Theisel, ``The state of the art in vortex extraction,'' \emph{Computer Graphics Forum}, vol.~37, no.~6, pp. 149--173, 2018.

\bibitem{Kasten2011Acceleration}
J.~Kasten, J.~Reininghaus, I.~Hotz, and H.~Hege, ``Two-dimensional time-dependent vortex regions based on the acceleration magnitude,'' \emph{{IEEE} Transactions on Visualization and Computer Graphics}, vol.~17, no.~12, pp. 2080--2087, 2011.

\bibitem{Bujack2020FlowSTAR}
R.~Bujack, L.~Yan, I.~Hotz, C.~Garth, and B.~Wang, ``State of the art in time-dependent flow topology: Interpreting physical meaningfulness through mathematical properties,'' \emph{Computer Graphics Forum}, vol.~39, no.~3, pp. 811--835, 2020.

\bibitem{Kasten2012VortexMerge}
J.~Kasten, I.~Hotz, B.~R. Noack, and H.~Hege, ``Vortex merge graphs in two-dimensional unsteady flow fields,'' in \emph{EuroVis 2012 - Short Papers}, M.~Meyer and T.~Weinkauf, Eds.\hskip 1em plus 0.5em minus 0.4em\relax Eurographics Association, 2012.

\bibitem{denby2022characterising}
L.~Denby, S.~J. B{\"o}ing, D.~J. Parker, A.~N. Ross, and S.~M. Tobias, ``Characterising the shape, size, and orientation of cloud-feeding coherent boundary-layer structures,'' \emph{Quarterly Journal of the Royal Meteorological Society}, vol. 148, no. 742, pp. 499--519, 2022.

\bibitem{couvreux2010resolved}
F.~Couvreux, F.~Hourdin, and C.~Rio, ``{Resolved versus parametrized boundary-layer plumes. Part I: A parametrization-oriented conditional sampling in large-eddy simulations},'' \emph{Boundary-layer meteorology}, vol. 134, pp. 441--458, 2010.

\bibitem{BinMasood2021b}
T.~B. Masood and I.~Hotz, ``Continuous histograms for anisotropy of 2d symmetric piece-wise linear tensor fields,'' in \emph{Anisotropy across fields and scales: Imaging, Geometry, and Astronomy.}, ser. Mathematics + Visualization.\hskip 1em plus 0.5em minus 0.4em\relax Springer, 2021, pp. 39--70.

\bibitem{Morozov2013b}
D.~Morozov, K.~Beketayev, and G.~H. Weber, ``Interleaving distance between merge trees,'' \emph{Discrete and Computational Geometry}, vol.~49, no.~52, pp. 22--45, 2013.

\end{thebibliography}

\vskip -2.2\baselineskip plus -1fil
\begin{IEEEbiography}[{\includegraphics[width=1in,height=1.25in,clip,keepaspectratio]{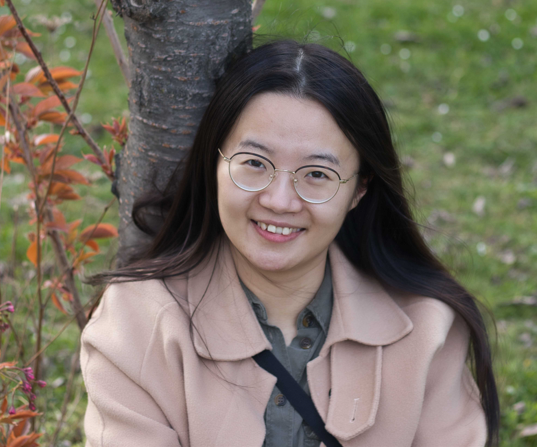}}]{Danhua Lei } is currently working towards the PhD degree in the Scientific Visualization group at Linköping University, Sweden. Before this, she received her master's degree in Scientific Computing from Heidelberg University, Germany. Her research interests are diverse and interdisciplinary, encompassing the fields of scientific visualization, medical visualization, and sparse representation.\end{IEEEbiography}

\vskip -2.2\baselineskip plus -1fil
\begin{IEEEbiography}[{\includegraphics[width=1in,height=1.25in,clip,keepaspectratio]{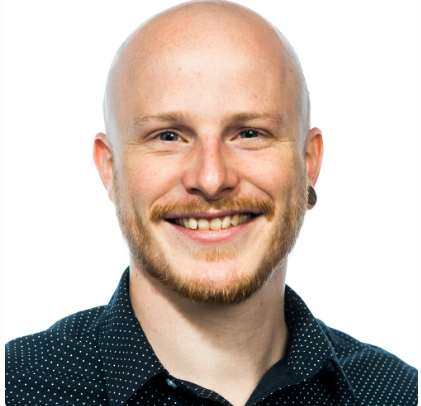}}]{Jochen Jankowai } is a doctoral student at Linköping University, Sweden, where he also obtained his master's degree in 2016. His research focuses specifically on tensor field exploration, feature space exploration, and analysis.\end{IEEEbiography}

\vskip -2.2\baselineskip plus -1fil
\begin{IEEEbiography}[{\includegraphics[width=1in,height=1.25in,clip,keepaspectratio]{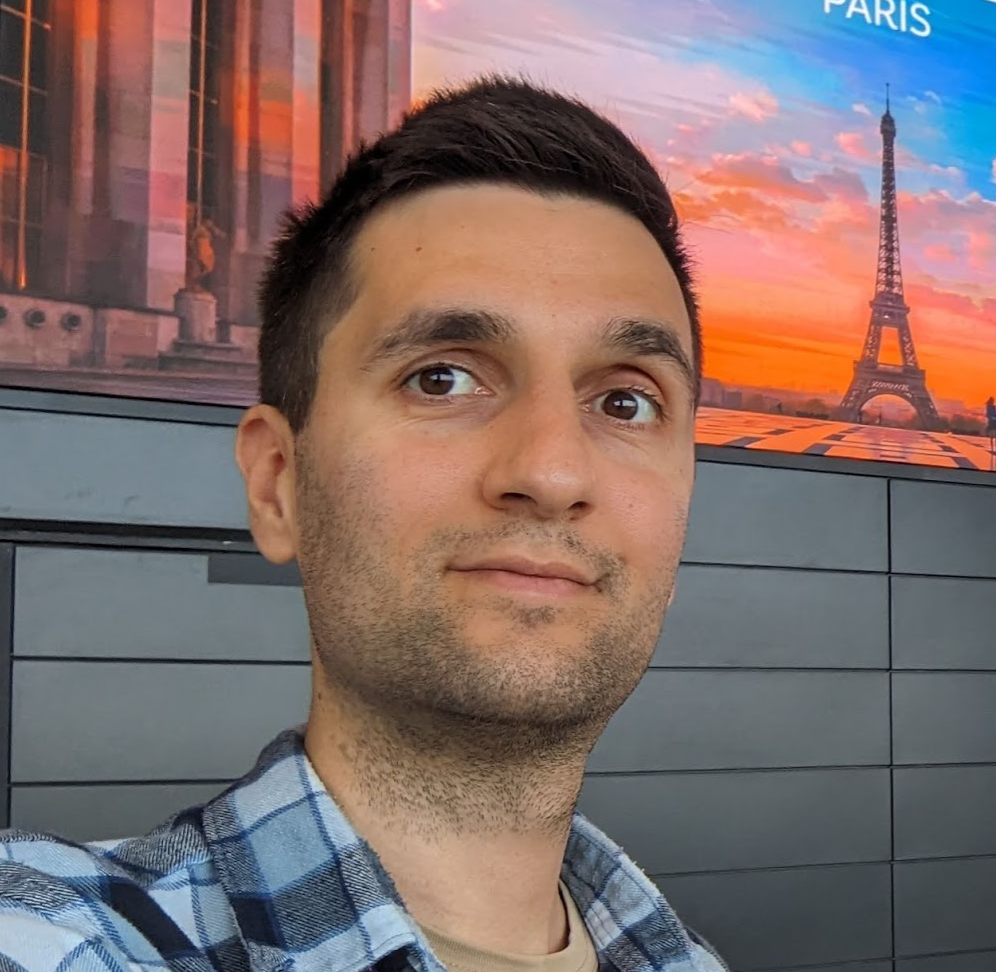}}]{Petar Hristov}
is a postdoctoral researcher at the scientific visualization group at Linköping University in Sweden. He received his PhD from the University of Leeds in 2022. His research interests include topological analysis and scientific visualization. This involves designing algorithms for topological data structures and working with scientists to analyze their data using topological methods.
\end{IEEEbiography}

\vskip -2.2\baselineskip plus -1fil
\begin{IEEEbiography}[{\includegraphics[width=1in,height=1.25in,clip,keepaspectratio]{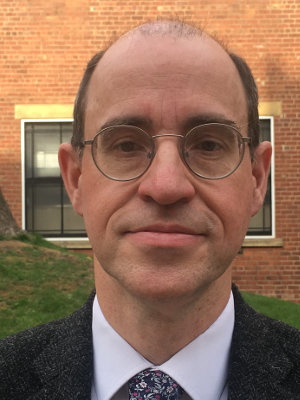}}]{Hamish Carr}
is a Professor in the School of Computing at the
  University of Leeds, having earned his PhD from the University of British
  Columbia in 2004. His research centers on computational topology and the
  mathematical foundations of scientific visualization, but also includes
  applications ranging from civil engineering to environmental sciences. He is
  a member of the IEEE, ACM and Eurographics, and currently serves as Corporate
  Secretary for Eurographics.
\end{IEEEbiography}

\vskip -2.2\baselineskip plus -1fil
\begin{IEEEbiography}[{\includegraphics[width=1in,height=1.25in,clip,keepaspectratio]{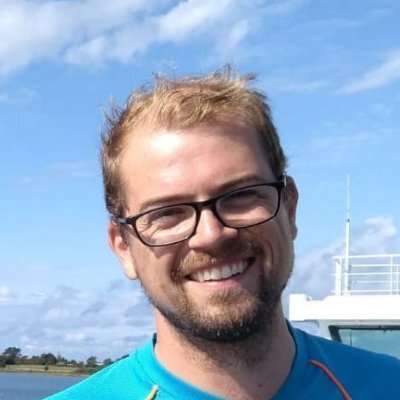}}]{Leif Denby}
is a senior research scientist at the Danish Meteorological Institute. He received his PhD from the University of Cambridge in 2015. His research focuses on using computational fluid dynamics and machine learning to study convective clouds.

\end{IEEEbiography}

\vskip -2.2\baselineskip plus -1fil
\begin{IEEEbiography}[{\includegraphics[width=1in,height=1.25in,clip,keepaspectratio]{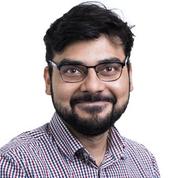}}]{Talha Bin Masood} is currently an assistant professor at Link\"{o}ping University in Sweden. 
He received his Ph.D. in Computer Science from the Indian Institute of  Science, Bangalore.  His research interests include scientific visualization, computational geometry, computational topology, and their applications to various scientific domains.\end{IEEEbiography}

\vskip -2.2\baselineskip plus -1fil
\begin{IEEEbiography}[{\includegraphics[width=1in,height=1.25in,clip,keepaspectratio]{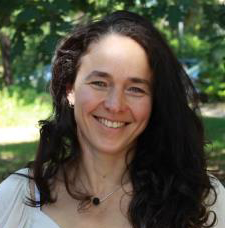}}]{Ingrid Hotz}
is currently a Professor in Scientific Visualization at the Link\"{o}ping University in Sweden. She received her Ph.D. degree from the Computer Science Department at the University of Kaiserslautern, Germany. Her research interests lie in data analysis and scientific visualization,  ranging from basic research questions to effective solutions to visualization problems in applications.  
\end{IEEEbiography}


\end{document}